\documentclass[11pt,a4paper]{article}
\usepackage[hyperref]{acl2021}
\usepackage{times}
\usepackage{latexsym}

\usepackage{microtype}

\aclfinalcopy 

\usepackage{graphicx}
\usepackage{booktabs}
\usepackage{enumitem}
\usepackage{multirow}
\usepackage{capt-of}
\usepackage{cleveref}

\definecolor{ao(english)}{rgb}{0.0, 0.5, 0.0}

\usepackage{enumitem}

\usepackage{xcolor}
\definecolor{kill}{HTML}{2ca25f}
\definecolor{arrest}{HTML}{8856a7}
\definecolor{force}{HTML}{e34a33}
\definecolor{all}{HTML}{43a2ca}
\definecolor{fail}{HTML}{dd1c77}

\def\killLabel{{\color{kill} \textsc{KILL}}} 
\def\arrestLabel{{\color{arrest} \textsc{ARREST}}} 
\def\forceLabel{{\color{force} \textsc{FORCE}}} 
\def\allLabel{{\color{all} \textsc{ANY ACTION}}} 
\def\failLabel{{\color{fail} \textsc{FAIL TO ACT}}} 

\title{Corpus-Level Evaluation for Event QA: \\ The IndiaPoliceEvents Corpus Covering the 2002 Gujarat Violence}

\author{Andrew Halterman$^*$ \\
  Massachusetts Institute of Technology \\
  \texttt{ahalt@mit.edu} \\\And
  Katherine A. Keith$^*$ \\
  University of Massachusetts Amherst  \\
  \texttt{kkeith@cs.umass.edu } \\ \AND
    Sheikh Muhammad Sarwar$^*$ \\
  University of Massachusetts Amherst \\
  \texttt{smsarwar@cs.umass.edu} \And
     Brendan O'Connor \\
  University of Massachusetts Amherst \\
  \texttt{brenocon@cs.umass.edu }
  }

\date{}

\begin{document}
\maketitle

\begin{abstract}
Automated event extraction in social science applications often requires corpus-level evaluations: for example, aggregating text predictions across metadata and unbiased estimates of recall. We combine corpus-level evaluation requirements with a real-world, social science setting and introduce the \textsc{IndiaPoliceEvents} corpus---all 21,391 sentences from  1,257 English-language \textit{Times of India} articles about events in the state of Gujarat during March 2002. Our trained annotators read and label every document for mentions of police activity events, allowing for unbiased recall evaluations. 
In contrast to other datasets with structured event representations, we gather annotations by posing natural questions, and evaluate off-the-shelf models for three different tasks: sentence classification, document ranking, and temporal aggregation of target events.
We present baseline results from zero-shot BERT-based models fine-tuned on natural language inference and passage retrieval tasks. Our novel corpus-level evaluations and annotation approach can guide creation of similar social-science-oriented resources in the future.

\end{abstract}

\begin{figure}[t]
\centering
\includegraphics[width=0.9\columnwidth]{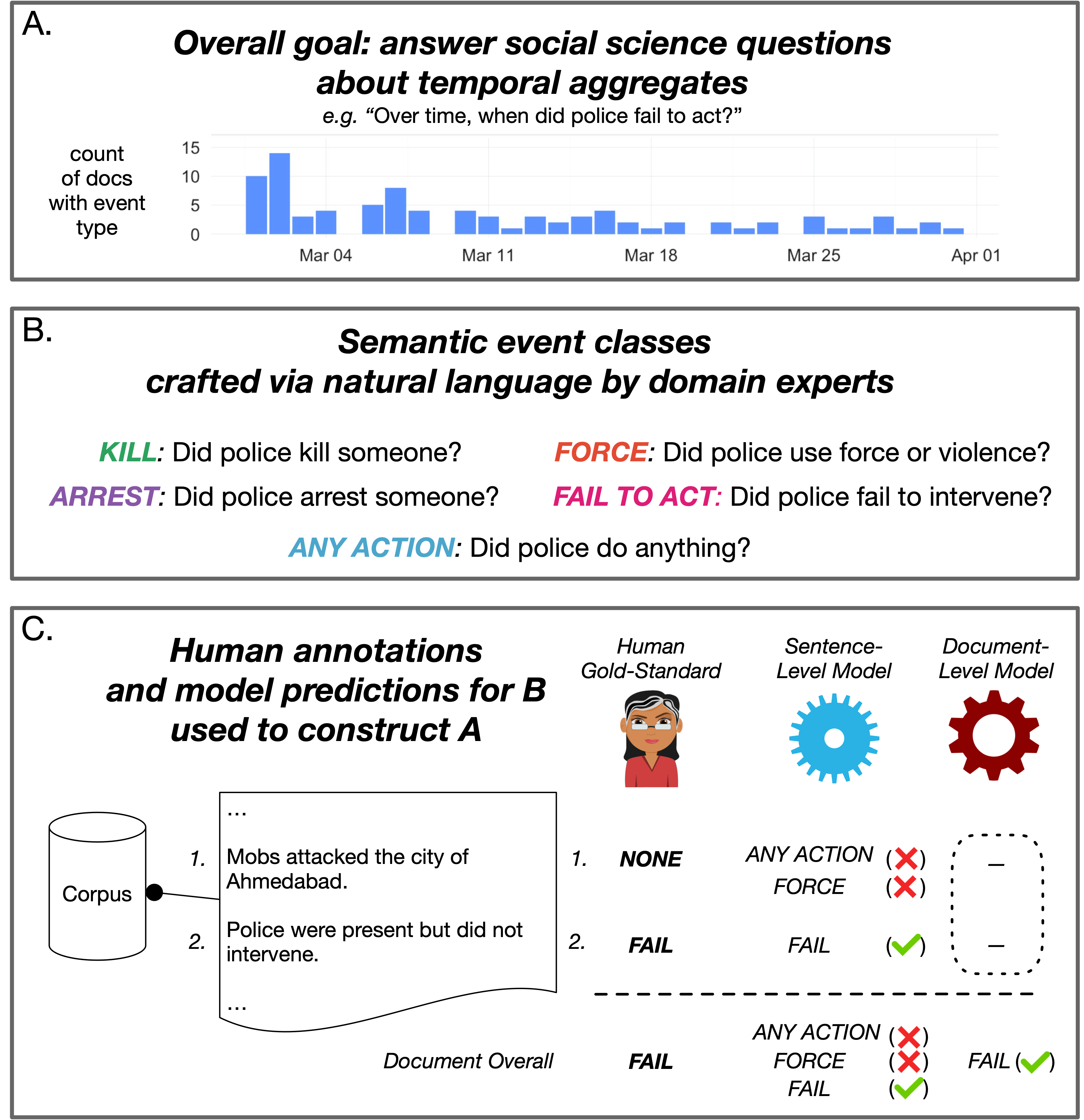}
\caption{Motivation (A-B) and procedures (B-C) for this paper: \textbf{A.} Social scientists often use text data to answer substantive questions about  temporal aggregates. \textbf{B.} To answer these questions, domain experts use natural language to define semantic event classes of interest. \textbf{C.} Our \textsc{IndiaPoliceEvents} dataset: Humans annotate \emph{every} sentence in the corpus in order to evaluate whether a system achieves full recall of relevant events. In production, computational models run B's queries to classify or rank sentences or documents, which are aggregated to answer A. \label{f:bigpic}
\vspace{-0.5cm}
} 
\end{figure}

\section{Introduction}\label{sec:intro}
Understanding the actions taken by political actors is at the heart of political science research: 
{\let\thefootnote\relax\footnotetext{$^*$ Indicates joint first-authorship.}}
How do actors respond to contested elections \citep{daxecker2019electoral}? How many people attend protests \citep{chenoweth2013unpacking}? Which religious groups are engaged in violence \citep{brathwaite2018measurement}? Why do some governments try to prevent anti-minority riots while others do not \citep{wilkinson2006votes}? In the absence of official records, social scientists often turn to news data to extract the actions of actors and surrounding events. These news-based event datasets are often constructed by hand, requiring large investments of time and money and limiting the number of researchers who can undertake data collection efforts. 

Automated extraction of political events and actors is already prominent in social science \citep{schrodt1994political,king2003automated,hanna2014developing,hammond2014using,icews2015,beieler2016generating,osorio2017supervised} and is increasingly promising given recent gains in information extraction (IE), the automatic conversion of unstructured text to structured datasets \cite{grishman1997information,mccallum2005information,grishman2019twenty}. While social scientists and IE researchers have overlapping interests in evaluating event extraction systems, social scientists have particular needs that have so far been under-addressed by the computer science IE research community. 

Figure~\ref{f:bigpic}A shows a common goal of social scientists: answering aggregate substantive questions from corpora such as, \textit{``Over time, when did police fail to act?''} which could be measured by, for example, the daily count of newspapers mentioning the event class over time. For these types of questions, social scientists predominantly want very high recall methods
because often the events of interest are sparse or their substantive conclusions depend on identifying \textit{every} event in a corpus.\footnote{In some studies, researchers rely on an assumption that events are missing at random, but others depend on knowing whether an event  occurred at least once.}

In contrast to this corpus-level focus, much of current IE research has focused on distinct subtasks such as entity linking, relation extraction, and coreference resolution.\footnote{The first five Message Understanding Conferences (MUC) required participants to submit complete systems to fill event templates; however, starting with MUC-6 and subsequent ACE and KBP tasks, information extraction was broken into distinct modules \cite{grishman1996message,grishman2019twenty}.} 
Furthermore, all widely used event datasets (e.g.\ ACE, FrameNet, ERE, or KBP; \citealt{aguilar2014comparison})
are typically curated at the \emph{ontology} level---attempting to cover a selected set of event types---but have little consideration of the \emph{corpus} level---annotated documents are not necessarily a 
substantively meaningful sample of the broader corpora from which they are drawn. We try to address these evaluation shortcomings in this paper. 

In addition to corpus-level recall, social scientists are often interested in using off-the-shelf models that are easily extensible to their domain questions. Fortunately, recent NLP research has seen a paradigm shift from structured semantic and event representations \citep{abend-rappoport-2017-state,aguilar2014comparison} which are limited by their predefined schemas,
to directly using natural language to encode semantic arguments (\textit{QA-SRL}; \citealt{he2015question,2016stanovskyAnnotating,fitzgerald2018qasrl,roit-etal-2020-controlled}) and events \citep{levy2017zero,liu2020event,du2020event}. In this paper, we also use natural language questions to annotate and model the event classes in our dataset, 
not only facilitating ease of annotation, but also allowing for the evaluation of 
zero-shot natural language inference and information retrieval models for the tasks.

To address these social science desiderata, we present the \textsc{IndiaPoliceEvents} corpus\footnote{Dataset, source code, and appendix are provided at \\
\url{http://slanglab.cs.umass.edu/IndiaPoliceEvents/} and 
\\ \url{https://github.com/slanglab/IndiaPoliceEvents}.}
which has the following useful properties: 
\begin{itemize}[noitemsep,leftmargin=*]
\itemsep0em
 \item \textbf{Social science relevance.} Our dataset consists of all 21,391 sentences from all 1,257 \textit{Times of India} articles about events in the state of Gujarat during March, 2002---a period that is of deep interest to political scientists due to widespread Hindu--Muslim violence  \citep{dhattiwala2012political,berenschot2012riot,basu2015violent}.  We focus on the actions of a single entity type,
 \emph{police}, because of extensive substantive research on police actions during the Gujarat violence \citep{varadarajan2002gujarat}. 
 Our choice of location, actors, and event types are motivated by \citet{wilkinson2006votes}---political science work which created a hand-coded event dataset from newspapers about communal violence events in India from 1950-1995.  
    \item \textbf{Corpus-level full-recall.} Unlike most previous event evaluation datasets, our annotators read \emph{every} document in our corpus (that match a loose spatiotemporal filter; \S\ref{sec:corpus}). This requires substantially more annotation work compared to a more targeted filter to select documents to annotate (e.g.~matching via keywords), but eliminates a potential source of evaluation bias
    compared to alternative document retrieval data collection approaches \cite{trec_toal_recall_2016},
    and allows for full-recall evaluation of end-to-end event extraction systems.  
     \item \textbf{Document-level context.} Our annotators read the context of an entire document to provide answers for each question on each sentence. We then aggregate these sentence-level answers to make document-level inferences. This allows us to accurately label sentences with anaphora or context-specific meaning. 
    \item \textbf{Natural language event specification and zero-shot model evaluation.} 
    In constructing our dataset, we gather annotations via a natural question-answer format because it allows for easily specifying constraints on arguments (e.g.\ \emph{police} being the agent). Additionally, it allows for specifying event predicates not covered within the ontologies of current structured semantic representations, or with additional hard-to-specify semantic phenomena---e.g.\ ``Did police fail to act?" or when political actors do \emph{not} take an action, which is very important to political scientists (e.g.~\citet{wilkinson2006votes}).
    This format also allows us to evaluate zero-shot  natural language inference and information retrieval models.
    \item \textbf{High-quality annotators who provide uncertainty explanations.} 
    We hire and train political science undergraduate students as annotators to ensure quality control, retraining annotators over a period of several months with training videos, two hour-long live meetings, and individual annotator feedback before producing our final dataset. Our annotators also provide free-text explanations for instances in which they are uncertain about the answer. These rationales are important given the recent attention to propagating annotator uncertainty in downstream NLP tasks \cite{dumitrache2018crowdsourcing,paun2018comparing,pavlick2019inherent,keith2020uncertainty} and social scientists' interest in quantifying uncertainty \citep{king1989unifying, wallach2018computational}.  
\end{itemize}

\noindent
In the remainder of this paper, we use our dataset for three levels of evaluation: sentence-level classification, ranking of documents to reduce manual reading time, and constructing temporal aggregates useful to social scientists (\S\ref{sec:task}). We describe in detail our annotation and dataset creation process (\S\ref{sec:data}), provide baseline models (\S\ref{sec:models}), and evaluate their performance on all three tasks (\S\ref{sec:evaluation}).  

\section{Related Work}
\label{sec:related-work}

\textbf{NLP and IR for police activity.}
Natural Language Processing (NLP) and Information Retrieval (IR) have been used for analysis of other police activity such as identifying victims of police fatalities from news articles \cite{keith2017identifying,nguyen-nguyen-2018-killed, sheikh_police_killing}; extracting eye-witness event types from Twitter including police activity and shootings \cite{doggett2016identifying}; detecting dialogue acts from police stops \cite{prabhakaran2018detecting}; and computational analysis of degree of respect in police officers' language \cite{voigt2017language}.

\textbf{Political event extraction.} Automated event extraction in social science is generally performed using dictionary methods and a set of substantively motivated event types and actor categories \citep{schrodt1994political,gerner2002conflict,beieler2016generating,boschee2016solutions,radford2016automated, brathwaite2018measurement,liang2018new}.
Other work uses supervised learning to infer events such as
conflict or cooperation \citep{beieler2016arxiv} and protests \citep{hanna2017mpeds}.
While some have attempted to induce event types without supervision \citep{oconnor2013learning,huang-etal-2016-liberal}, most social science applications of event extraction require substantial human input either through constructing keyword lists, or annotating texts to train classifiers.


\textbf{Recall-focused IR.} 
TREC's total-recall track \cite{trec_toal_recall_2016} is inspired by real-world recall-focused applications from law, medicine, and oversight \cite{graham_mcdonald_active_learning}. However, the track's datasets are not typically focused on events and assume documents are collected through interacting with a system. 
Other work has focused on methods for truncating ranked lists that minimize the risk of viewing non-relevant documents \cite{arampatzis2009stop,lien2019assumption}, but this line of work does not evaluate on semantic retrieval of event classes. 

\section{Three Levels of Tasks}
\label{sec:task}
In order to answer substantive social science questions, for example \textit{``Does variation in party control of state government affect whether police failed to intervene in ethnic conflict?''} \cite{wilkinson2006votes}, social scientists often need to gather counts of events (e.g.~``police failed to intervene'') from text when official government records are lacking. Ideally, a social scientist could use automatic information extraction methods \cite{cowie1996information,mccallum2005information,grishman2019twenty} to transform unstructured text into a structured database that would be useful in a quantitative analysis. Yet, even state-of-the-art information extraction systems often give less than perfect accuracy, so social scientists must still manually analyze large portions of their corpus in order to extract events of interest. This quantitative research process motivates the following three tasks which our dataset can be used to evaluate:

\textbf{Task 1: Sentence classification.} Although social science corpora typically consist of \textit{documents}, it would be useful for a system to classify \textit{sentences} that contain events of interest.\footnote{This is closely related to extracting ``explanation representations'' \cite{thayaparan2020survey}, ``supporting facts'' \cite{yang2018hotpotqa} or ``evidence sentences'' \cite{wang2019evidence} in the machine reading comprehension literature.} Highlighting relevant sentences could, for semi-automated systems, reduce a social scientist's reading time, and, for fully-automated systems, provide sentence-level evidence of the automated method's \emph{validity}, a crucial aspect of research in text-as-data \cite{grimmer2013text} and the broader social sciences \cite{drost2011validity}.
\textsc{IndiaPoliceEvents} allows for evaluation of sentence-level precision, recall, and F1 (\S\ref{ss:eval-task1}).  

\textbf{Task 2: Document ranking.} For semi-automated systems, social scientists must navigate the tradeoff between recall and manual reading time. Social scientists may rely on IR methods which present ranked lists of relevant documents \cite{baeza1999modern,schutze2008introduction}. However, our informal interviews with social scientists suggest they want to know at what point they have read enough documents to achieve very high (95--100\%) recall. In creating \textsc{IndiaPoliceEvents}, annotators read every single sentence in a corpus which allows for full evaluations of average precision and our newly proposed metric: the proportion of the corpus that would have to be read to achieve Recall=$X$ (PropRead@RecallX) (\S\ref{ss:eval-task2}).\footnote{We do not address the problem of estimating recall when gold-standard labels are only known for the subset of documents read so far, but \textsc{IndiaPoliceEvents} could be used to evaluate that task in future work.}

\textbf{Task 3: Substantive temporal aggregates.} 
For social scientists, NLP and IR methods are used in service of answering substantive questions from text. In addressing our running example \textit{``Did differences in party control of state government affect whether police failed to intervene in ethnic conflict?''} a social scientist could measure how many news articles\footnote{Count of news articles are often used in social science as a proxy for the true measure of the event, e.g. \citet{nielsen2013rewarding, chadefaux2014early}.} discuss ``police failing to intervene'' each day for a given temporal span. In this setting, it would be helpful to know if changes in model performance at the sentence or document level resulted in significant differences at this aggregate level.
We design \textsc{IndiaPoliceEvents} with the capability of evaluating these meaningful corpus-level temporal aggregates, such as the mean absolute error and Spearman rank correlation coefficient between per-day event counts of computational models and ground truth annotations (\S\ref{ss:eval-task3}).

\section{Annotations and Dataset}
\label{sec:data}


\subsection{Corpus selection}  \label{sec:corpus}
We curate our corpus with a substantively motivated specification: it is restricted to a single authoritative news source, over a defined span of time, with articles that mention one of two locations involved in or related to the 2002 Gujarat violence.

From the website of \textit{Times of India}, an English language newspaper of record in India, we first download all news articles published in March 2002.\footnote{\S\ref{sec:ethics} discusses copyright issues.} During this period, widespread communal violence occurred in India, following the death of 59 Hindu pilgrims in a train fire in the state of Gujarat. In the subsequent months, reprisal attacks were directed at mostly Muslim victims across the state \citep{hrw2002orders,subramanian2007political}.  In creating our annotations, we specifically focus on the actions of police during these events,
since a large body of evidence points to the importance of police intervention and non-intervention in quelling or permitting ethnic violence \citep{hrw2002orders,wilkinson2006votes,subramanian2007political}. We focus on the first month of the violence in order to fit within our annotation budget. This month saw the greatest levels of violence, though violence continued for a period of months afterward.

Our final corpus consists of the subset of scraped documents published in March 2002 that include either the name of the state (\textit{Gujarat}) or a city related to the beginning of violence (\textit{Ayodhya}).\footnote{Selecting documents using location-based keywords is a standard first step in political science text analysis \citep{mueller2017reading}. 
This filters to 18\% of the total articles in March 2002. The precipitating event for the March 2002 violence was the burning of a train of pilgrims returning from Ayodhya.} Selecting on geographical and temporal metadata is a high recall way to filter the corpus without biasing the dataset by filtering to topic or event-related keywords, thus giving a better view of the true recall of an event extraction method.
\begin{table}[t]
    \centering
    \resizebox{0.98\linewidth}{!}{
    \begin{tabular}{lrrrr}
    \toprule
         Event Class & \multicolumn{2}{c}{Pos.~Sents.} & \multicolumn{2}{c}{Pos.~Docs.} \\
         \cmidrule(lr){1-1} \cmidrule(lr){2-3} \cmidrule(lr){4-5}
         \killLabel & 96 &(0.45\%) & 50 & (3.98\%)\\
         \arrestLabel& 299 &(1.40\%) & 128 &(10.17\%) \\
         \failLabel & 207 &(0.97\%) & 114 &(9.05\%)\\
         \forceLabel & 222 &(1.04\%) & 90 &(7.15\%) \\
         \allLabel & 2,073 &(9.69\%)	& 457 &(36.24\%) \\
         \bottomrule
    \end{tabular}
    }
    \caption{\textsc{IndiaPoliceEvents} number and percentage of positive sentences (sents.)\ and documents (docs.)\ after the adjudication round. In total, the dataset contains 21,391 sentences and 1,257 documents.
     \label{tab:positives}}
\end{table}

\subsection{Annotations via natural language}\label{ss:label-descriptions}
\begin{figure}
    \centering
    \includegraphics[width=0.6\columnwidth]{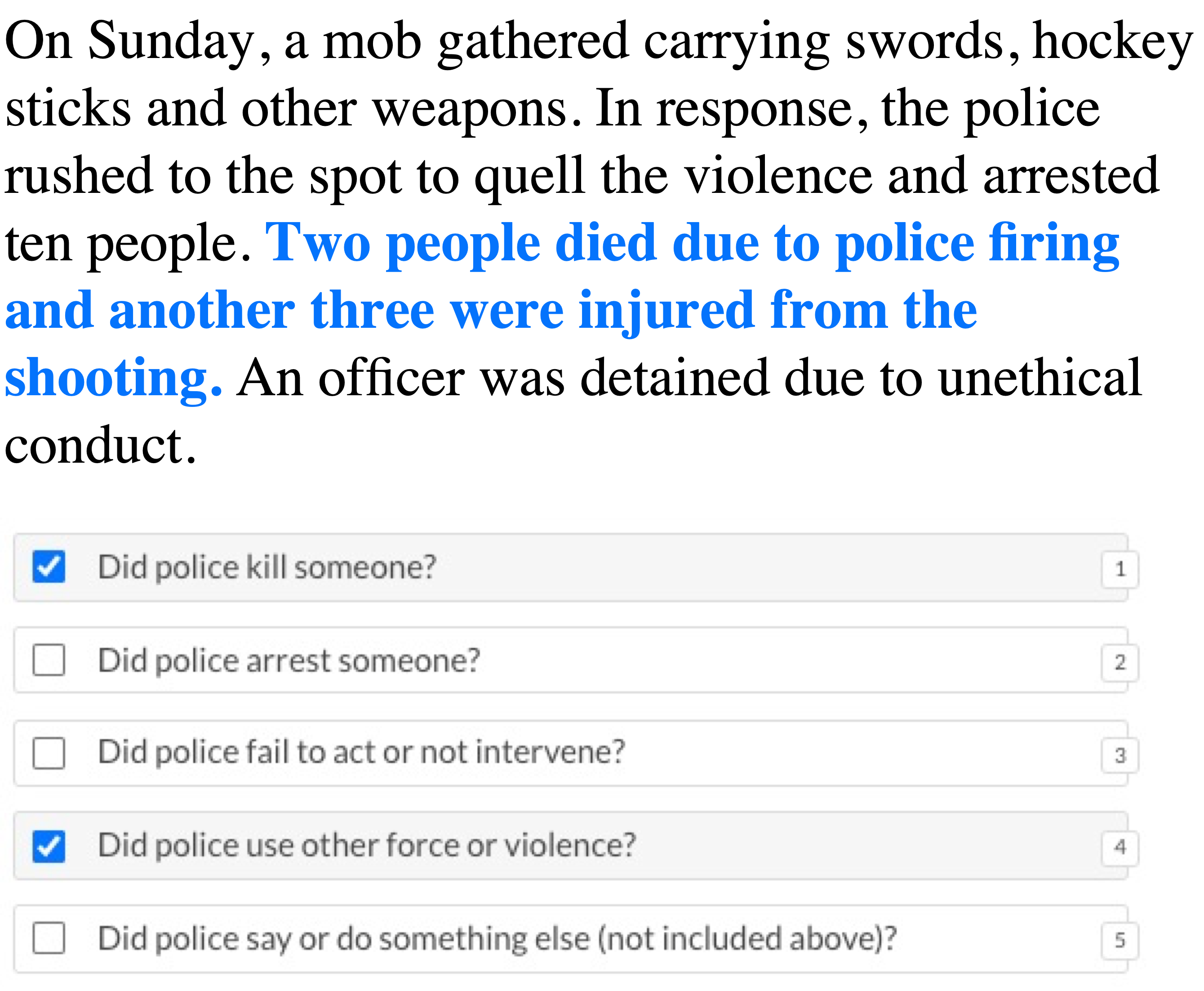}
    \caption{We present annotators with a highlighted sentence (blue) and its document context. Their task is to click a check-mark for the event-focused questions for which there is a positive answer in the highlighted sentence.  
    \label{fig:fig1}}
\end{figure}

To collect annotations, we give annotators an entire document for context, and then ask them \textit{natural language questions} about semantic event classes anchored on the actions of police for each sentence in that document:

\begin{itemize}[noitemsep,leftmargin=*]
\itemsep0em
    \item 
    \killLabel:
    ``Did police kill someone?'' Lethal police violence is an important subject for social scientists \citep{subramanian2007political}. Example sentence: \textit{``In Vadodara, one person was killed in police firing on a mob in the Fatehganj area."}
    \item 
    \arrestLabel:
    ``Did police arrest someone?" Knowing when and where police made arrests and who was arrested is an important part of understanding police response to communal violence. Example sentence: \textit{``Police officials said nearly 2,537 people have so far been rounded up in the state.''}
    \item
    \failLabel: 
    ``Did police fail to intervene?" In the 2002 Gujarat violence, police were often accused of failing to prevent violence or allowing it to happen. Knowing when police were present but did not act is important for understanding the extent of this phenomenon and its potential causes \citep{wilkinson2006votes}. Example sentence: \textit{``The news items [...] suggest inaction by the police force [...] to deal with this situation."}

    \item 
    \forceLabel: ``Did police use force or violence?'' Political scientists are interested not only when police kill but the level of force they use. Example sentence: \emph{``Trouble broke out in Halad [...] where the police had to open fire at a violent mob."}
    \item \allLabel: ``Did police do anything?'' We collect annotations on all police activities, so that social scientists could, in the future, label more fine-grained event classes. Example sentence: \textit{``In the heart of the city's Golwad area, the army is maintaining a vigil over mounting tension following [...]"}
\end{itemize}
Figure~\ref{fig:fig1} shows the interface annotators see.\footnote{
While the first three classes each correspond to a single annotation question, 
we create \forceLabel~and \allLabel~by taking the union of several different questions posed to annotators,
which made it easier for annotators to distinguish between different subtypes.
\forceLabel~is the union of ``Did police kill someone?'' and ``Did police use other force or violence?''.
\allLabel~is the union of four questions:
``Did police kill someone?'', 
``Did police arrest someone?", 
``Did police use other force or violence?'',
and ``Did police do or say something else (not included above)?''.
\label{footnote:anno_class_unions}
} See Appendix \S\ref{sec:annotation} for exact annotation instructions and per-question agreement rates.

Following the guidelines of \citet{pustejovsky2012natural}, we first assign each document to two annotators and then follow with an \emph{adjudication round} in which items with disagreement are given to an additional annotator to resolve and create the gold standard.
For annotators, we select undergraduate students majoring in political science (as opposed to crowdworkers) in order to approximate the domain expertise of social scientists.\footnote{
Our annotation protocol (no.\ 2238) was reviewed as exempt by the University of Massachusetts Amherst's IRB office. Annotators were paid \$25 per training session and a lump sum for document annotations;
we expected this to exceed \$14 USD per hour based on a generous (conservatively high) estimate of completion time.
All annotators reported their work time was less than this estimate.}
We initially recruited and selected 12 students. After a pilot study and two rounds of training, in which we provided individual feedback to annotators via email, we selected 8 final annotators based on their performance. Each student annotated around 330 documents ($\sim$5,500 sentences) using the interface described in the Appendix, Figure~\ref{f:interface}.

Table~\ref{tab:positives} shows the prevalence of the event classes after the adjudication round. Note that some of the classes are relatively rare: of all documents, only roughly 4\% have \killLabel~and 7\% have \forceLabel. Our annotators had fairly high inner-annotator agreement for \killLabel~and \arrestLabel, with Krippendorff's alpha values of 0.75 and 0.71 respectively. Other questions, such as \failLabel~and ``Did police use other force?'' had lower agreement ($\alpha < 0.4$), indicating more difficulty and ambiguity. Full agreement rates are show in Appendix, Table~\ref{t:agreement}.


\subsection{Annotation uncertainty explanations}
We also collect free-text \emph{annotation uncertainty explanations} in order to analyze instances that annotators found difficult or ambiguous. For each sentence presented to annotators, we ask ``If you found this example difficult or ambiguous please explain why'' and ask them to provide a short written response in a provided text-box. This follows recent work that that has emphasized the importance of annotator disagreement not necessarily always as error in annotation but instead as ambiguity that is inherent to natural language and a potential useful signal for downstream analyses \cite{dumitrache2018crowdsourcing,paun2018comparing,pavlick2019inherent,keith2020uncertainty}.

 Annotators remarked on several types of text they were uncertain about: agents of actions who were not explicitly mentioned but implicitly police, named entities whose status as police is ambiguous, confusion about what precisely constitutes an ``arrest'', and confusion arising from the lack of specific cultural knowledge (e.g., around the Indian crowd-control tactic of ``lathi charging''). In the appendix, see Table~\ref{t:interesting} for examples and Table~\ref{t:free_text} for a categorization of free text responses.




\section{Baseline Models}
\label{sec:models}

We test several baseline models, all requiring no annotation (and thus most realistic for the social science use case), and assess their performance on \textsc{IndiaPoliceEvents}.

\textbf{Keyword matching.} 
Boolean keyword queries are a very common social science approach to document classification (e.g.\ \citet{nielsen2013rewarding,chadefaux2014early,d2014separating,baker2016measuring}), since they are simple, transparent, and widely supported in user software.
We use conjunctive normal form rules, where inferring an event class for a sentence requires matching any term from a \emph{police} keyword list (including both common nouns and names of major police and security institutions), as well as an event keyword.
To construct the keyword lists,
a domain expert coauthor first manually generates a list of seed keywords for the semantic categories \emph{police}, \emph{kill}, \emph{arrest}, \emph{intervention}, and \emph{force}. To address lexical coverage,  we then expand the keywords through word2vec \citep{mikolov2013distributed} nearest neighbors, filtered to semantically equivalent words by the domain expert.\footnote{We train word2vec on every article in the \textit{Times of India} from 2002 (the same corpus as our dataset, 69,000 articles) plus another 100,000 articles from \textit{The Hindu}, another English-language newspaper in India. We inspect each keyword's 20 nearest neighbors with highest cosine similarity.} This process is repeated using WordNet synonym sets for lookup \cite{miller1995wordnet}, resulting in 217 keywords total; see appendix (\S\ref{sec:appdx-keywords}) for details.


\textbf{RoBERTa+MNLI.} 
Given two input sentences, a premise and hypothesis, the task of natural language inference (NLI) is to predict whether the premise \textit{entails} or \textit{contradicts} the hypothesis or does neither (\textit{neutral}) \cite{bowman2015large,williams2018broad}.
Previous work has shown promise of NLI transfer learning for events:
\citet{sheikh_event_retrieval} show Sentence-BERT embeddings \cite{reimers2019sentence} learned from NLI data are effective on ACE-like event retrieval; \citet{clark2019boolq} find for \emph{BoolQ}, their dataset of naturally occurring boolean questions, that transfer learning from NLI data is more effective than transferring from QA or paraphrase data.  We follow \citet{clark2019boolq}'s example and use ``off-the-shelf'' RoBERTa \cite{liu2019roberta} fine-tuned on the MNLI corpus \cite{williams2018broad}. 
The model takes a sentence and a declarative form of an event class question as input (\S\ref{sec:apdx-declarative}), and we use its predicted probability of entailment as the probability of the event class. For document ranking, we create a document score by taking the maximum predicted probability over sentences. Future experiments could vary the amount of text (sentence vs.~passage~vs.~document) used as input to the model.

\textbf{BM25+RM3.}
Weighted term matching between a query and document
is a strong competitor to neural ranking methods
\citep{craswell2020overview,jimmy_lin_classification}, 
via, for example,
BM25 scoring with RM3 query expansion \cite{lavrenko_language_models}. 
With the Anserini BM25 implementation \cite{anserini},
we set $k_1$ = 0.9 and $b$ = 0.4, and conduct RM3 expansion of the query to terms found in the top $k=10$ BM25-retrieved documents, following \citet{jimmy_lin_classification}'s hyperparameter settings.
As the input query, this set of models uses the natural language questions described in \S\ref{ss:label-descriptions}. Appendix Table \ref{t:bm} contains full results.

\textbf{ELECTRA+MS MARCO.} Fine-tuned BERT \citep{devlin2019bert} and other large-scale language models have been used extensively for document ranking in information retrieval (IR) \cite{zhan2020repbert,zhang-etal-2020-little, dai_document_ranking, sean_document_ranking}. We use a competitive off-the-shelf model that uses the ELECTRA variant of BERT \cite{clark2020electra} fine-tuned on MS MARCO \cite{reimers2019sentence}.
MS MARCO is a large-scale reading comprehension dataset in which questions are sampled from anonymized web queries, and answers to the queries are generated by crowdworkers \cite{nguyen2016ms}. It was used in the 2019 TREC Deep Learning track on document and passage retrieval \cite{craswell2020overview}. We use \citet{reimers2019sentence}'s pretrained ELECTRA+MS MARCO model.
Inputs are \textsc{IndiaPoliceEvents} passages consisting of three-sentence sliding windows with stride of one sentence and queries are the event class questions described in \S\ref{ss:label-descriptions}. Following related work \citep{zhan2020repbert}, we take the maximum score of all passages as the document score.

\section{Results}
\label{sec:evaluation}
We report the performance of the baseline models (\S\ref{sec:models}) on our three tasks in Table~\ref{t:results-big-table} and in  Figure~\ref{fig:combo_eval}. 

\setlength{\tabcolsep}{5pt}   

\newcommand{\fscore}{{\footnotesize F1 $\uparrow$}}
\newcommand{\ap}{{\footnotesize AP $\uparrow$}}
\newcommand{\pr}{{\footnotesize PR $\downarrow$}}
\newcommand{\mae}{{\footnotesize MAE $\downarrow$}}
\newcommand{\myrho}{{\footnotesize $\rho\uparrow$}}

\begin{figure*}[h!] 
\noindent\begin{minipage}{\linewidth}
\centering

 \begin{tabular}{l cc cccccc cc}
        & \multicolumn{2}{c}{\textit{Task 1: Sent.\ Cls.}} 
        & \multicolumn{6}{c}{\textit{Task 2: Document Ranking}}
        & \multicolumn{2}{c}{\textit{Task 3: Temp. Aggs.}}\\

        \cmidrule(lr){2-3} \cmidrule(lr){4-9} \cmidrule(lr) {10-11}
        
        & \multirow{1}{*}{Keyw.} 
        & \multirow{1}{*}{\shortstack[l]{R+MNLI}}
        & \multicolumn{2}{c}{
          \multirow{1}{*}{BM25}} 
        & \multicolumn{2}{c}{
          \multirow{1}{*}{\shortstack[l]{E+MSM}} }
        & \multicolumn{2}{c}{
          \multirow{1}{*}{\shortstack[l]{R+MNLI}}}

        & Keyw.
        & R+MNLI
          \\
        Event Class & \fscore & \fscore & \ap& \pr & \ap & \pr & \ap& \pr & \myrho & \myrho  \\
        \cmidrule(r){1-1} 
        \cmidrule(lr){2-2} 
        \cmidrule(lr){3-3}
        \cmidrule(lr){4-5}
        \cmidrule(lr){6-7}
        \cmidrule(lr){8-9}
        \cmidrule(lr){10-10}
        \cmidrule(lr){11-11}

{\small \killLabel}  &0.50&\textbf{0.74}&0.30&0.29&0.65&0.27&\textbf{0.96}&\textbf{0.05}&     0.70 & \textbf{0.78}\\
{\small \arrestLabel}  &0.48&\textbf{0.62}&0.68&0.36&0.72&0.67&\textbf{0.91}&\textbf{0.17} &  0.71 & \textbf{0.85}\\
{\small \failLabel}  &0.05&\textbf{0.48}&0.27&0.77&0.36&0.87&\textbf{0.63}&\textbf{0.76} &     0.42 & \textbf{0.60}\\
{\small \forceLabel} &\textbf{0.65}&0.62&0.24&0.43&0.64&0.45&\textbf{0.90}&\textbf{0.20}&     \textbf{0.89} & 0.86\\
{\small \allLabel} &\textbf{0.67}&0.57&0.53&0.85&0.83&0.88&\textbf{0.89}&\textbf{0.62}&       0.86 & \textbf{0.90}\\
\bottomrule
  \end{tabular}
\captionof{table}{ Evaluation of two classification methods (Keyw., R+MNLI) and three ranking models (BM25, E+MSM, R+MNLI) for \textsc{IndiaPoliceEvents}'s three tasks. Bolded numbers indicate the model that performs best on each metric and event class. \textbf{Task 1} evaluates sentence-level F1 for sentence-level keyword matching (Keyw.) and RoBERTa fine-tuned on MNLI (R+MNLI) \cite{liu2019roberta}. \textbf{Task 2} evaluates average precision (AP) and proportion of the corpus needed to be read in order to achieve 95\% recall (PR, or PropRead@Recall95) for ranking models BM25 \cite{anserini}, off-the-shelf ELECTRA language model fine-tuned on MS MARCO (E+MSM; \cite{reimers2019sentence}), as well as R+MNLI's probabilistic output. \textbf{Task 3} evaluates Spearman's rank correlation coefficient ($\rho$) between predicted versus gold-standard counts of documents with the relevant event, for each day in March 2002.
   For each metric, we indicate whether a higher ($\uparrow$) or lower ($\downarrow$) score is better. \label{t:results-big-table}}

\bigskip{}
\centering
\includegraphics[width=0.8\textwidth]{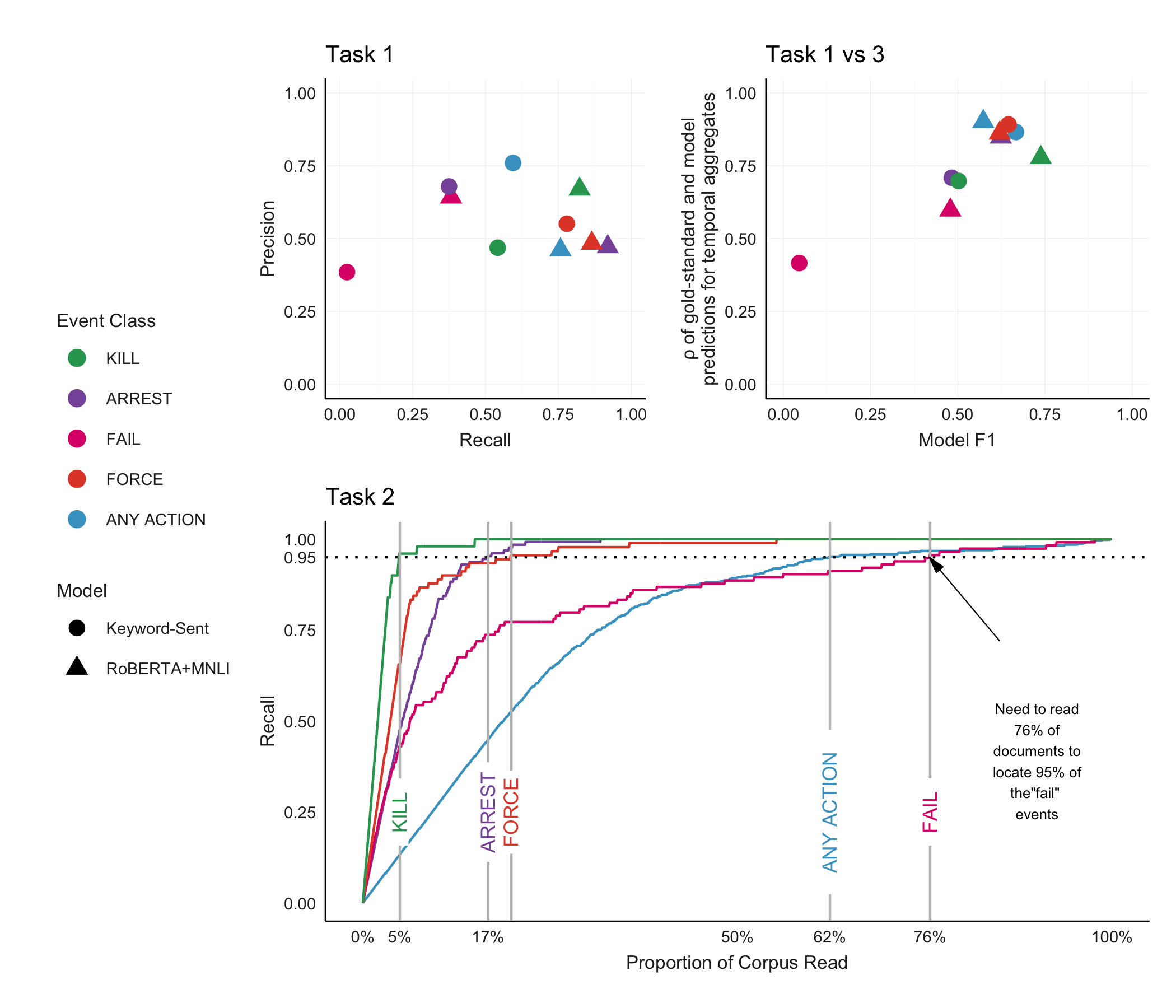}
\captionof{figure}{Keyword and RoBERTA+MNLI performance on three metrics. 
\textbf{(Task 1)} Precision and recall at the sentence level for two models on each semantic event class. 
\textbf{(Task 1 vs 3)} Sentence-level model F1 scores (x-axis) versus Spearman's $\rho$ of hand-annotated gold-standard and model predictions for temporal aggregates (y-axis).  
\textbf{(Task 2)} For each class under RoBERTa MNLI, the gain curves \cite{trec_toal_recall_2016}, marking PropRead@Recall95:
What percentage of the ranked corpus would a researcher need to read in order to find 95\% of each event classes' mentions? 
\label{fig:combo_eval}}

\end{minipage} 
\end{figure*}

\subsection{Task 1: Sentence classification.}\label{ss:eval-task1}  
For sentence classification,\footnote{We do not evaluate sentences with less than 5 tokens as many of these sentences are due to sentence segmentation errors. After this filtering, the number of remaining sentences we evaluate on is 18,645.} 
Table~\ref{t:results-big-table} shows that the keyword matching method slightly outperforms RoBERTa+MNLI on F1 
for \allLabel~and \forceLabel, which  
we suspect is due to the keyword method having better access to synonyms of ``police” (e.g.~``jawan'', ``RPF'') particular to the \textit{Times of India} via its \textit{word2vec} expansion. However, RoBERTa+MNLI achieves a higher F1 score on \killLabel, \arrestLabel, and \failLabel.
We need further controlled experiments to understand how the concreteness of the event class, importance of identifying events' agents, and formulation of the query (e.g.~``Did police use force or violence?'' vs.~``Were police violent?'') affect the results of contextualized language models. 
Table~\ref{t:results-big-table} also shows poor performance of our keyword matching method on \failLabel\ (F1=0.05); however, a large-scale contextual language model seems to be able to better distinguish the semantics of the event class (F1=0.48). 
The Task 1 plot in Figure~\ref{fig:combo_eval} shows that across all labels, RoBERTa+MNLI has higher recall than the keyword method for every event class. If social scientists plan to use these sentence classification methods in a semi-automated fashion (as we suggest in \S\ref{sec:intro}), selecting models like RoBERTa+MNLI that achieve higher recall may be important. 

\subsection{Task 2: Document ranking.}\label{ss:eval-task2}
For Task 2, we report average precision and a new
metric---the proportion of documents that would have to be read to achieve recall equal to $X$ (\textit{PropRead@RecallX}). 
We use $X=0.95$ because social scientists typically use 95\% cutoffs for significance and sampling error.\footnote{We note that 5\% recall error is not equivalent to a 5\% sampling error. In practice, researchers are more concerned with whether data is missing at random.} We leave to future work estimating recall on a corpus without ground truth. 
Table~\ref{t:results-big-table} shows that RoBERTa+MNLI outperforms both BM25 and ELECTRA+MS MARCO on both average precision and PropRead@Recall95 across all event classes. We hypothesize this is because natural language inference is a task that is much more aligned with the semantic-oriented precision at which we want to rank documents. In contrast, the MS MARCO dataset is constructed for a much higher level information need, and documents that are ``relevant” could potentially not entail the semantic event class of interest.
As Figure~\ref{fig:combo_eval} shows, if a social scientist was presented with a ranked list of documents from RoBERTa+MNLI, they would only have to read 5\% of the entire corpus to achieve 95\% recall on \killLabel.
RoBERTa+MNLI also does well on \arrestLabel~and \forceLabel~with 0.17 and 0.20 PropRead@Recall95 respectively.  There is consistently more difficulty across all models for \allLabel~and \failLabel. We speculate this is because \allLabel~is the class with the greatest prevalence, and thus is more difficult to achieve higher recall.


\begin{figure}
    \centering
    \includegraphics[width=0.95\columnwidth]{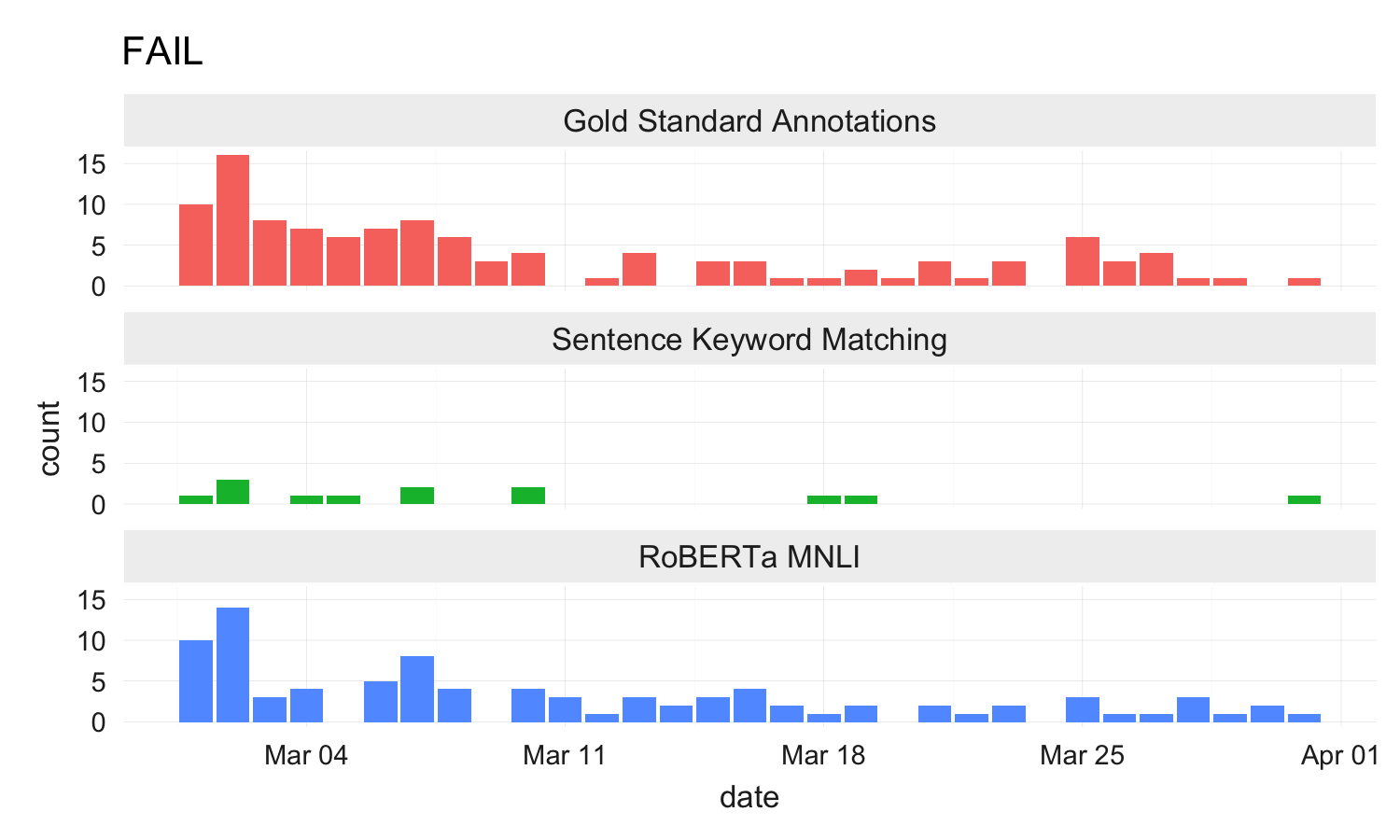}
    \caption{Number of documents per day containing a ``police failing to act/standing by'' event, comparing outputs from \textbf{(Top)} human annotations \textbf{(Middle)} keyword matching (Spearman's $\rho$ = 0.51, comparing with gold standard) and \textbf{(Bottom)} RoBERTa finetuned on MNLI ($\rho$ = 0.60).}
    \label{fig:aggregate}
\end{figure}

\subsection{Task 3: Temporal Aggregates.}\label{ss:eval-task3}
Figure~\ref{fig:aggregate} compares the outputs of three systems on \failLabel: gold-standard human annotations, keyword matching, and RoBERTa+MNLI. For this event class, both automated methods under-count the number of events marked by human annotators. In contrast, the automated techniques tend to \textit{over}count other event types (see Appendix, Figure~\ref{fig:all_aggs} for plots of the other event classes). While the the overall temporal trend is broadly consistent across the three methods, the decreased accuracy of the automated methods could lead to attenuation bias if they were used as input to statistical models. A qualitative examination of the extracted events also reveals the need for future work in temporal linking models: most of the events after March 25 are describing events from earlier in March that were being reported in the context of investigations into the violence. Table~\ref{t:results-big-table} shows that for all event classes except for  \forceLabel~RoBERTa+MNLI has a higher Spearman's $\rho$\footnote{See Table~\ref{tab:appdx-mae} for mean absolute error scores.} between the predicted versus gold-standard document counts. The Task 1 vs.~3 plot in Figure~\ref{fig:combo_eval} shows an approximately linear relationship between the F1 scores of sentence-level models and Spearman's $\rho$, suggesting there is promise that NLP research focused on sentence-level models could be of use to social scientists who care about corpus-level evaluation. 

\subsection{Qualitative error analysis.} 
We manually analyze the false positives and false negatives of our best-performing baseline model, RoBERTa+MNLI. Some false positives are due to lexical semantic misunderstandings:  
the model often mistakes ``shot'' for \killLabel, and assigns high probability to negative \forceLabel~sentences such as, \textit{“The police escorting the vehicle fired into the air and dispersed the mob.''} 
The model also has difficulty identifying the police as agents: for example, it assigns high probability to the negative \killLabel~sentences: 
\textit{“[…] scores of people have been killed in rural Gujarat due to police failure […],”} and \textit{“Police said that two persons had been killed in Vijaynagar […]”}.
Other errors are due to \emph{hypotheticals}: the model assigns high probability to the negative \killLabel~sentences \textit{“He alleged BJP's hand in the murder of […]”} and \textit{“Achar claims she was an eye-witness to police complicity in the violence.”} 
Many of the model’s false negatives are due to necessary multi-sentence context (which RoBERTa+MNLI does not have as it only takes single sentences as input). For instance, the model assigns low probability to the positive \killLabel~sentence \textit{“Four persons have been killed and five are injured.”} and \forceLabel~sentence \textit{``One person was injured and rushed to the SSG hospital"}; if one reads the proceeding context of both of these sentences it is clear that police are the agents of the actions. 


\section{Discussion and Future Work} 
The dataset, tasks, and evaluations we present in this work are driven by the needs of social scientists: we assess the performance of zero-shot models on metrics important to applied researchers, including recall against a fully annotated corpus and performance at temporally aggregated levels. We find cause for optimism for social scientists using BERT-style pre-trained models on their tasks. These models could potentially be used in place of social scientists' existing keyword-based classifiers, although we caution accuracy is far from perfect and applied researchers will need to extensively validate model outputs. Even with imperfect classification accuracy, we believe these zero-shot models show promise for decreasing human annotation effort by reducing the proportion of the corpus read to achieve a specific recall level (the metric we call PropRead@RecallX).

Future work can extend our dataset creation process to new semantic event classes, such as protests, communal violence itself, and other forms of participation in political and social activity. Additional annotated datasets could allow researchers to generalize the performance of zero-shot language models to new domains and event classes. 
Finally, tasks such as temporal and geographic linking, event de-duplication and coreference, and identifying hypothetical events are unsolved but are major obstacles for applied social scientists working with automatically extracted events. 

\section{Acknowledgments}
For helpful comments, we're grateful to Aidan Milliff, the UMass NLP group, and anonymous ACL reviewers. This work was funded by a Kaggle Open Data Science grant, and additionally:
\emph{AH:} a National Science Foundation Graduate Research Fellowship;
\emph{KK:} a Bloomberg Data Science PhD Fellowship;
\emph{SS:} the Center for Intelligent Information Retrieval (CIIR);
\emph{BO:} National Science Foundation IIS-1845576 and IIS-1814955.

\section{Ethical Considerations and Broader Impact}  \label{sec:ethics}

To ensure the replicability of our work and to further research into event extraction systems for social science research, we are making the text of the news articles available to researchers alongside our annotations.
While all articles were obtained from a public website without login credentials,
the applicability of copyright restrictions is relevant to address.

We believe the research benefits and the limited harms to the copyright holders justify this use,
due to the four criteria considered in the fair use doctrine in U.S.\ copyright law \cite{copyright_office}:
(1) the non-commercial, nonprofit educational purpose of our use of the text,
(2) the factual nature of the news reports, 
(3) the limited substitutability of our dataset for the original news site,\footnote{We do not republish the texts as consumer-accessible webpages, but instead are only contained within a JSON structured format.}
and (4) our expectation that our limited corpus will not harm the market for readers of the news site.
 
The issue of copyright status within NLP-oriented corpora is of increasing interest.
\newcite{sag2019copyright} argues machine learning uses of text is non-expressive and therefore falls under fair use, and \newcite{geiger2018copyright}
study the issue in the context of proposals for E.U.\ law.
\newcite{bandy2021addressing} investigate BooksCorpus, a previously poorly documented corpus widely used for training language models, finding it contains large amounts of copyrighted work, 
highlighting how current data curation practices in machine learning (and adjacent) communities need improvement \cite{paullada2020data,Jo2020Lessons}.

We also acknowledge the sensitivities around this period of violence in India. Its significance motivates computational work to enable more effective study of it and related episodes, but our news-derived data on its own, in the absence of deeper qualitative work, does not permit us to draw new substantive conclusions about the causes and consequences of the violence in Gujarat in 2002. We defer to the large scholarly and journalistic literature on the violence; see references in \S\ref{sec:intro} and \S\ref{sec:data}.

\bibliographystyle{acl_natbib}
\bibliography{acl2021}

\clearpage

\appendix

\setcounter{table}{0}
\renewcommand{\thetable}{A\arabic{table}}
\setcounter{figure}{0}
\renewcommand{\thefigure}{A\arabic{figure}}


\section*{Appendix}

\section{Annotation Details} \label{sec:annotation}

We provide details on our annotation process here, including the semantic event class definitions we provided to annotators, the per-class agreement statistics, statistics on the time it took to annotate, and further qualitative analysis of the annotations.

All results reported in this appendix correspond to responses to the annotation questions (shown in \Cref{t:agreement}), which are slightly different from the main semantic event classes reported in the main paper, as described in \Cref{t:agreement}'s caption and \Cref{footnote:anno_class_unions}.

\subsection{Annotator Instructions}

\begin{figure*}
    \centering
    \includegraphics[width=\columnwidth]{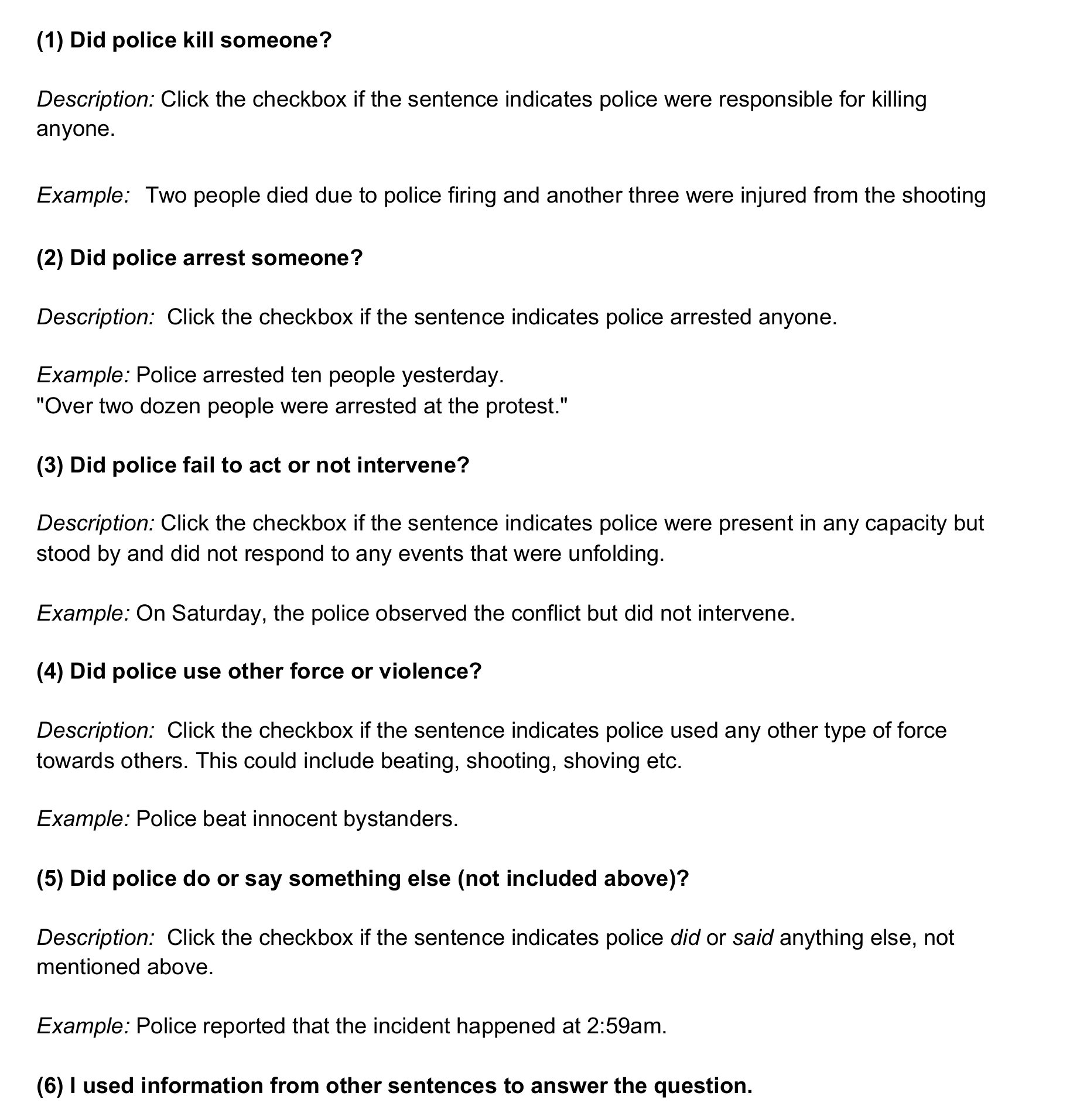}
    \includegraphics[width=\columnwidth]{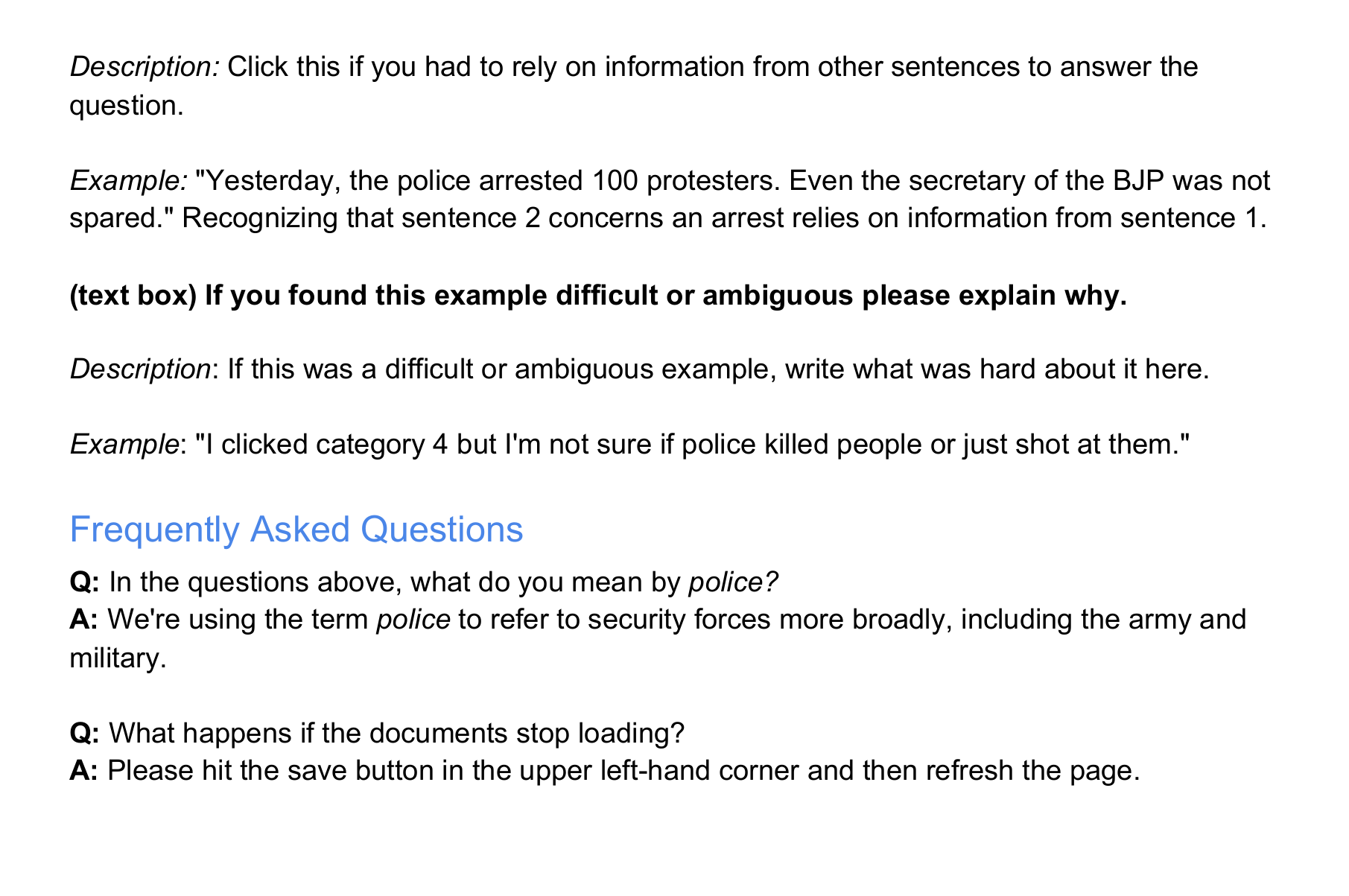}
    \caption{Instructions provided to annotators providing additional guidance on how to interpret the questions, giving an example positive sentence, and clarifying other issues that arose in training.
     \label{fig:appdx-annot-instr}}
\end{figure*}

To train our annotators, we provided them with a the question for each semantic event class, a short description to clarify the question, and an example positive sentence (Figure \ref{fig:appdx-annot-instr}). We met with the annotators as a group to talk through the document and then gave them a training round with documents we had previously annotated. Based on that training round, we added frequently asked questions to the instructions document, provided individual feedback to annotators, and then began the production annotation process on our corpus. 

\subsection{Annotation Interface}

\begin{figure*}[t]
\centering
\includegraphics[width=0.95\textwidth]{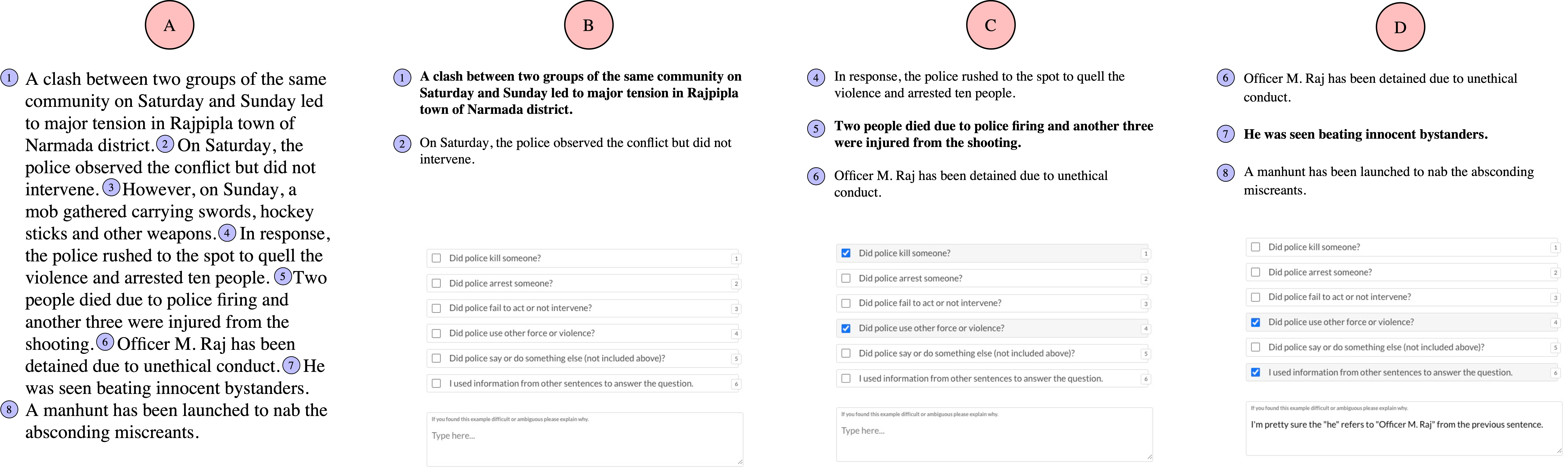}
\caption{
Illustration of the dataset annotation interface given an example document. In practice, annotators view and label \emph{all} sentences in the document, but this figure highlights \emph{three} informative sentence examples. (A) An example document with post-hoc numbering of sentences. (B) The user sees a bold sentence its context and then is asked a series of yes/no questions about the bold sentence. For this example, the annotators do not check any boxes (the answer to all the questions is \emph{no}). (C) For this bold sentence, annotators check the boxes (\emph{yes} answers) for the questions ``Did police kill someone?'' and ``Did police use other force or violence.'' (D) For this bold sentence, annotators check the box for ``Did police use other force or violence?'', select ``I used information from other sentences to answer the question,'' and provide a free-text explanation for why they thought the example was difficult. \label{f:interface}}
\end{figure*}

Figure \ref{f:interface} shows a stylized version of the custom interface we built using the Prodigy annotation tool \citep{montani2018prodigy}. Annotators are presented with an entire document, with sentences sequentially highlighted. For each highlighted question, they are asked each of the questions. If the sentence contains a positive answer to the question(s), they select the corresponding box(es) and advance to the next sentence. 

\subsection{Multi-sentence labels}

We record whether annotators report using information from other sentences in the document to annotate the current sentence. Specifically, we provide a checkbox in the interface with the label ``I used information from other sentences to answer the question". We collected this information in order to understand the number of sentences that could be classified on their own and how many needed broader document context. We caution that we left the interpretation of the sentence up to each annotator and did not train them or compare their usage of this label as we did with other labels. We do not use the labels in our analysis but provide them in our dataset to potentially help future research.

\subsection{Annotator Agreements}

We calculated inter-annotator agreement, both raw agreement and Krippendorff's alpha, for all annotators on our corpus (Table \ref{t:agreement}). Because the event classes are rare in our corpus, we prefer Krippendorff's alpha over raw agreement, which is inflated by the large number of zeros in our data. After half of the documents were annotated, we calculated agreement to check for annotators with high disagreement. We found one annotator with high disagreement on the \killLabel~class and provided updated instructions for them. We used the final agreement rates to select the three annotators with the highest agreement rates with the full set of annotators to serve as our adjudicators in the final round of annotations.

\begin{table*}[t]
\centering
\resizebox{0.98\linewidth}{!}{
  \begin{tabular}{l r r r r r}
  \toprule
Question & Agr. (all) & Agr. (1+)  & Krip. (all) & Krip. (1+)  & Support (1+) \\
\toprule
(1) ``Did police kill someone?" & 0.998 & 0.984 & 0.753 & 0.751  & 108  \\
(2) ``Did police arrest someone?" & 0.993 & 0.949 & 0.734 & 0.710  & 328  \\
(3) ``Did police use other force or violence?'' & 0.995 & 0.961 & 0.704 & 0.686  & 227  \\
(4) ``Did police fail to act or not intervene?" & 0.988 & 0.907 & 0.418 & 0.377  & 339  \\
(5) ``Did police say or do anything else?'' & 0.942 & 0.555 & 0.587 & 0.086 & 2142  \\
  \toprule
  \end{tabular}
  }
 \caption{
 Sentence-level agreement, Krippendorff, and support for each question answered by annotators. ``All'' refers to all 20,527 annotated sentences in the corpus and ``(1+)'' refers to a subset of the corpus that excludes sentences that both annotators agree do not have police actions. Questions 1, 2, and 4 map to \killLabel, \arrestLabel, and \failLabel, respectively;
\forceLabel\ is defined as (\emph{1 OR 3}), and \allLabel\ is (\emph{1 OR 2 OR 3 OR 5}),
as described in \Cref{footnote:anno_class_unions}.
 \label{t:agreement}
 }
\end{table*}

\subsection{Annotation Timing} 

\begin{figure}
    \centering
    \includegraphics[width=\columnwidth]{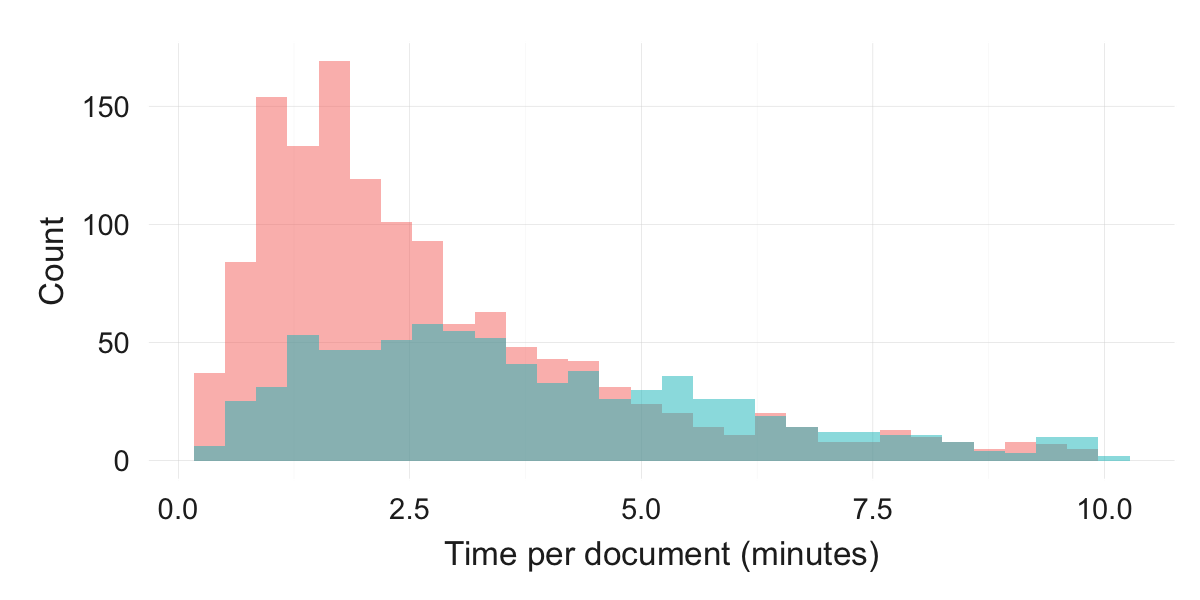}
    \caption{Time to annotate a document in minutes. The median time to annotate one document is 3.0 minutes, with documents that contain no police events (red) taking less time than documents with police activity (blue). Not shown are 366 out of 2,514 documents taking longer than 10 minutes.  Median document length is 15 sentences, min length=2, 25\% = 9, 75\%=22, max=98.}
    \label{fig:timing}
\end{figure}

Figure \ref{fig:timing} shows the distribution of the time that annotators took to annotate each document.

\section{Properties of Annotated Data}

\subsection{Event locality}
\begin{figure}
    \centering
    \includegraphics[width=0.95\columnwidth]{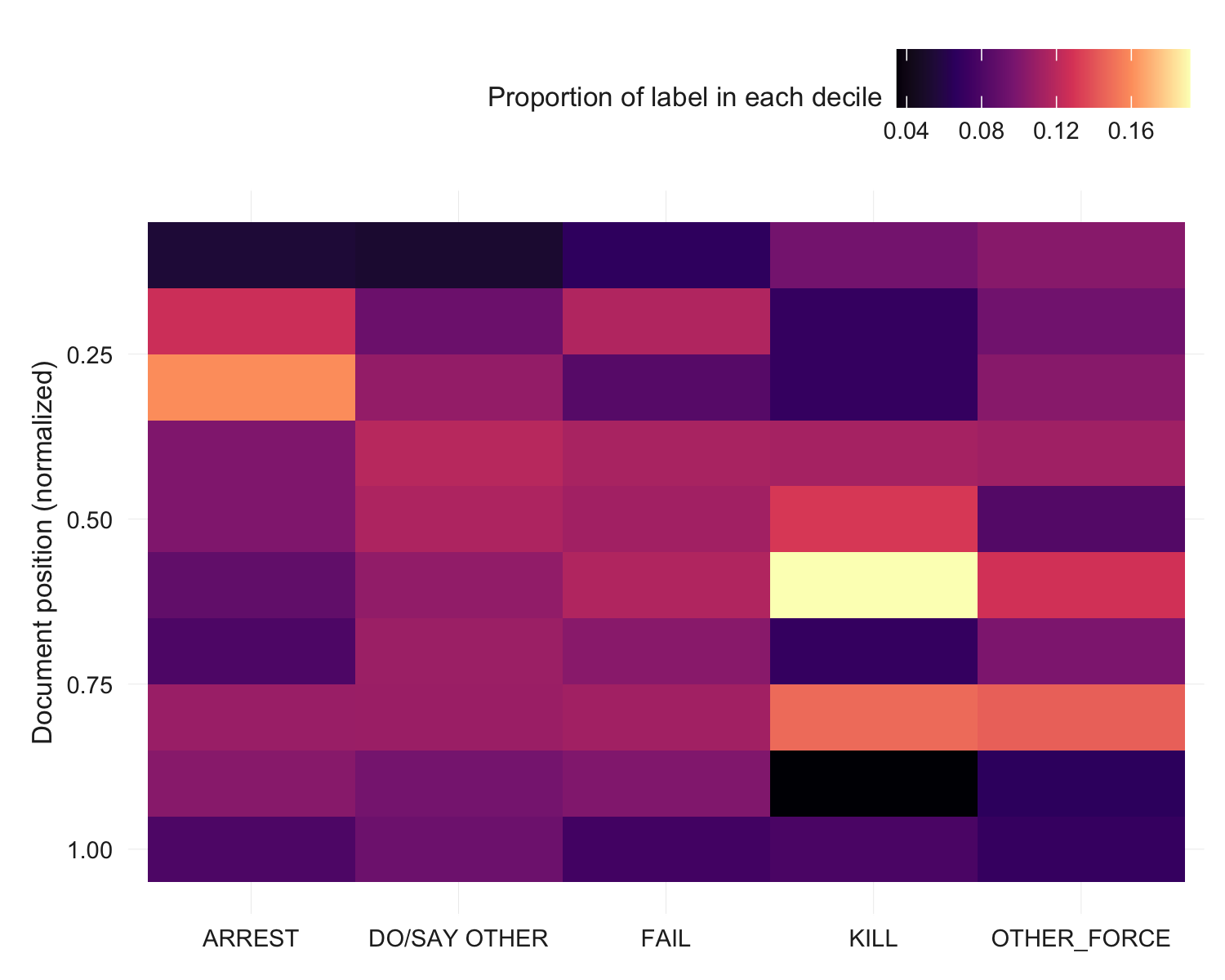}
    \caption{The location within a document of answers to each of the questions.  Questions are  often answered in the second half of the document.}
    \label{fig:label_locs}
\end{figure}

Figure \ref{fig:label_locs} shows the the label density for each event class in different sections in a document. 
To compute the label density, we partition the sentences in each document into ten equal and ordered sections, where the first section indicates the earliest location in a document. Then we compute the number of positive event in those sections. 
Figure \ref{fig:label_locs} shows that except for \arrestLabel, label density is not high in the initial sections of a document. In information retrieval and news summarization, typically the first $k$ tokens are assumed a good  approximation for document representation \cite{dai_document_ranking}, which our dataset seems to present contradictory evidence. 

\subsection{Analysis of Free-Text Explanations}

We analyzed the free text explanations given by annotators, grouping them into non-exclusive categorizes. The most common categories of annotator explanations are shown in Table \ref{t:free_text}.

\begin{table}[]
    \centering
    \begin{tabular}{l r}
    \toprule
    \textbf{Author-assigned category} & \textbf{Count} \\
    \midrule
No explicit agent &  63 \\
Agent may not be police & 57 \\
Police are mentioned but not agents & 37 \\
Hypothetical or future events &  35 \\
Failing to act vs. acting and failing &  30 \\
True ambiguity in language & 27 \\
Ambiguity in ``arrest'' & 26 \\
\midrule
Total free text explanations & 311 \\
\midrule
Total sentences with explanations & 299 \\
Total sentences with police activity$^*$ & 2,783\\ 
Total sentences in corpus & 21,391 \\
\toprule
    \end{tabular}
    \caption{Top categories of free-text explanations by annotators. Text explanations can be assigned to multiple categories.
    $^*$Here, ``total sentences with police activity'' means at least one annotator noted it had police activity. 
    \label{t:free_text}
    }
\end{table}

\subsection{Interesting Examples}

Table \ref{t:interesting} shows a selection of sentences from our corpus that illustrate challenging annotation decisions for our annotators or interesting ambiguity in the sentences.

\begin{table*}[t]
\renewcommand{\arraystretch}{1.25}
\centering
  \begin{tabular}{p{7.5cm}p{7.5cm}}
  \toprule
  Sentence from \textsc{IndiaPoliceEvents} & Author comment    \\
  \toprule
  At one point, the men in khaki seemed to have outnumbered the vhp workers 
  & Keyword matching would have missed ``men in khaki'' as a reference to police. \\
  \midrule
  The police have rounded up 1,740 VHP activists headed for Ayodhya. & Annotators flagged ambiguity in ``rounded up'' vs. ``arrested'' \\
  \midrule
  One of them who was on duty on December 16 even recollected how he and another colleague had to burst the tear gas shell themselves as the constables deliberately looked the other way. & Annotators flagged this sentence as two separate police agencies acted and failed to act.\\
  \midrule
  Police on Friday lathicharged ram sevaks who attempted to rush towards the make-shift temple in the disputed site here giving some anxious moments to security forces. & ``Lathi charges'' are an Indian riot control tactic that our United States-based annotators were not familiar with. \\
  \midrule
  Meanwhile the district administration has tightened the security in and around the temple city. & Many annotators flagged sentences where police are implicitly the agents. \\
  \bottomrule 
  \end{tabular}
  \caption{Example sentences illustrating several of the challenges of annotating the documents or in applying existing models. We provide our own commentary on why the sentences are difficult. 
  \label{t:interesting}  
  }
\end{table*}

\section{Modeling Details}

\subsection{Declarative versions of questions}\label{sec:apdx-declarative}
We use the following declarative versions of event class labels as input to RoBERTa+MNLI: 
\begin{itemize}
    \item \killLabel: ``Police killed someone."
    \item \arrestLabel: ``Police arrested someone.''
    \item \failLabel: ``Police failed to intervene."
    \item \forceLabel: ``Police used violence.''
    \item \allLabel: ``Police did something.''
\end{itemize}

\subsection{Keyword Approach}\label{sec:appdx-keywords}

We report here the terms used in the keyword matching. These terms were generated using subject matter expertise and expanded using WordNet and a custom word2vec model trained on the complete set of \textit{Times of India} articles from 2002 and 100,000 additional articles from the Indian newspaper \textit{The Hindu}. The expanded set was filtered using subject matter expertise. We report the keywords on the following categories: 

\emph{Police:} police, policemen, cop, cops, constables, constables, jawan, jawans, grp , cid , rpf , stf , bsf , dcp , dsp , ssp , sho , cisf , dgp.  

\emph{Kill}: kill, kills, killed, killing, lynch, lynched, lynching, annihilate, annihilating, annihilated, annihilates, drown, drowning, drowned, drowns, massacre, massacring, massacred, massacres, slaughter, slaughtering, slaughtered, slaughterers, butcher, butchering, butchered, butchers, poison, poisoning, poisoned, poisons, exterminate, exterminating, exterminated, exterminates, strangle, strangling, strangled, strangles, impale, impaling, impaled, impales, murder, murdering, murdered, murders, execute, executing, executed, executes. \\ 

\emph{Arrest}: arrest, arresting, arrested, arrests, nab, nabbing, nabbed, nabs, detain, detaining, detained, detains, book, booking, booked, books, chargesheet, chargesheeting, chargesheeted, chargesheets, apprehend, apprehending, apprehended, apprehends, seize, seizing, seized, seizes, collar, collaring, collared, collars. \\ 

\emph{Intervention}: intervene, intervening, intervened, intervenes, intervention, interfere, interfering, interfered, interferes, stand by, standing by, stood by, stands by, abstain, abstaining, abstained, abstains. \\ 

\emph{Force}: fire, firing, fired, fires, stone-pelt, stone-pelting, stone-pelted, stone-pelts, pelt stones, pelting stones, pelts stones, pelted stones, beat, beating, beaten, beats, whip, whipping, whipped, whips, bash, bashing, bashed, bashes, choke, choking, choked, chokes, wound, wounding, wounded, wounds, strong-arm, strong-arming, strong-armed, strong-arms, pistol-whip, pistol-whipping, pistol-whipped, pistol-whips, lash, lashing, lashed, lashes, trounce, trouncing, trounced, trounces, cane, caning, caned, canes, thrash, thrashing, thrashed, thrashes, clobber, clobbering, clobbered, clobbers, spank, spanking, spanked, spanks, paddle, paddling, paddled, paddles, hit, hitting, hits, whack, whacking, whacked, whacks, pummel, pummelling, pummeling, pummelled, pummeled, pummeled, pummels, club, clubbing, clubbed, clubs, shoot, shooting, shot, shoots, suffocate, suffocating, suffocated, suffocates, beat, beating, beaten, beats. \\ 

The keyword-matching method uses the following rules to classify a sentence or document:
\begin{itemize}
    \item \killLabel: If a \emph{police} keyword \emph{AND} a \emph{kill} keyword appear in the same piece of text, classify it as a positive. 
    \item \arrestLabel: If a \emph{police} keyword \emph{AND} an \emph{arrest} keyword appear in the same piece of text, classify it as a positive. 
    \item \failLabel: If a \emph{police} keyword \emph{AND} an \emph{intervention} keyword appear in the same piece of text, classify it as a positive. (This is a very simple rule-based method and we leave to future work to develop a keyword-based method that more adequately captures the \emph{not} semantics of ``did not intervene.'')
    \item \forceLabel: If a \emph{police} keyword \emph{AND} an \emph{force} keyword appear in the same piece of text, classify it as a positive. 
    \item \allLabel: If a \emph{police} keyword appears in a piece of text, classify it as a positive. 
\end{itemize}

\subsection{RoBERTa+MNLI.}
We use the pretrained model from \url{https://github.com/pytorch/fairseq/blob/master/examples/roberta/README.md#pre-trained-models} 

\subsection{ELECTRA+MS MARCO.}

We use the pretrained model from the \texttt{sentence-transformers} package \url{https://github.com/UKPLab/sentence-transformers/tree/master/examples/applications/information-retrieval#pre-trained-cross-encoders-re-ranker}.

For IR, there are two standard architectures for scoring passages and queries: \emph{cross-encoders} in which the architecture performs full attention over the pair and \emph{bi-encoders} in which the passage and query are each mapped independently into a dense vector space \cite{luan2020sparse}. We chose a model with a cross-encoder architecture since these have been shown to consistently have higher performance \cite{thakur2020augmented}.





\section{Results}

This section provides additional results beyond those included in the main paper, including results for document-level models, variants on the BM25 model, mean absolute error results to complement the Spearman correlations presented in the main paper, and the temporally aggregated results for all of the semantic event classes.

\subsection{Document Level F1}

To complement the sentence-level F1 metrics in the main paper, we present the document-level metrics in Table \ref{t:results-doc}.

\begin{table*}[t]
  \centering
      \begin{tabular}{lrrrr}
      \toprule
&& \multicolumn{3}{c}{\textit{Document Level F1}} \\ 
&Prop. Pos. & Keyword-Sent & Keyword-Doc & RoBERTa+MNLI\\ \toprule
\allLabel&0.36&0.82&0.82&0.63\\
\arrestLabel&0.10&0.68&0.50&0.63\\
\forceLabel&0.07&0.70&0.48&0.59\\
\killLabel&0.04&0.61&0.40&0.77\\
\failLabel&0.09&0.13&0.24&0.60\\
\bottomrule
  \end{tabular}
  \caption{Document level metrics for keyword matching and RoBERTa+MNLI model. ``Keyword-sent'' aggregates results from the sentence keyword matcher to the document level using an at least one threshold.\label{t:results-doc}}
\end{table*}

\subsection{BM25 and Variants}

In addition to the standard BM25 model reported in the paper, we tested several variants, including automatic term expansion using RM3 and manual term expansion using the same keywords from our keyword method. The results are shown in Table \ref{t:bm}.

\begin{table*}[t]
  \centering
      \begin{tabular}{lrrrr}
      \toprule
& \multicolumn{4}{c}
{\textit{Ranking Metrics}: Ave. Precision ($\uparrow$), PropRead@Recall95 ($\downarrow$)}
\\ 
& BM25 & BM25+expn & BM25+RM3 & BM25+RM3+expn  \\ \toprule
\allLabel&0.53,~0.85&0.76,~0.85&0.50,~0.83&0.65,~0.81\\
\arrestLabel&0.68,~0.36&0.63,~0.21&0.58,~0.44&0.45,~0.43\\
\forceLabel&0.24,~0.43&0.52,~0.44&0.26,~0.42&0.67,~0.50\\
\killLabel&0.30,~0.29&0.22,~0.36&0.28,~0.33&0.36,~0.49\\
\failLabel&0.27,~0.77&0.24,~0.69&0.25,~0.72&0.21,~0.66\\
\bottomrule
  \end{tabular}
  \caption{Comparisons of BM25 and its variants. Here, \textit{expn} means we use the manually-curated expanded keywords and append these to the original query as input into the model. \label{t:bm}}
\end{table*}

\begin{table*}[h!]
    \centering
    \begin{tabular}{l r r}
    \toprule
    Event Class & Keyword (MAE) & RoBERTa+MNLI (MAE) \\
    \midrule
     \killLabel  & 2.16 & 0.867 \\
    \arrestLabel  & 2.10 & 4.45 \\
    \failLabel  & 3.25 & 1.48 \\
    \forceLabel  & 7.22 & 3.23 \\
    \allLabel  & 37.87 & 15.42 \\
    \bottomrule
    \end{tabular}
    \caption{Mean absolute error (MAE) between the human gold-standard annotations and the keyword-matching and RoBERTa+MNLI models. 
    \label{tab:appdx-mae}}
\end{table*}

\subsection{Spearman and MAE results}

In the main paper, we report Spearman correlations between the daily count of gold standard events identified by our annotators. In Table \ref{tab:appdx-mae} we also report the mean absolute error in daily event counts between our two models for each event class. We prefer Spearman correlations over MAE because the correlation is normalized between -1 and 1, while MAE tends to be higher for high-prevelance event classes.

\subsection{Temporal Aggregates for All Event Classes}

\begin{figure*}
    \centering
    \includegraphics[width=\columnwidth]{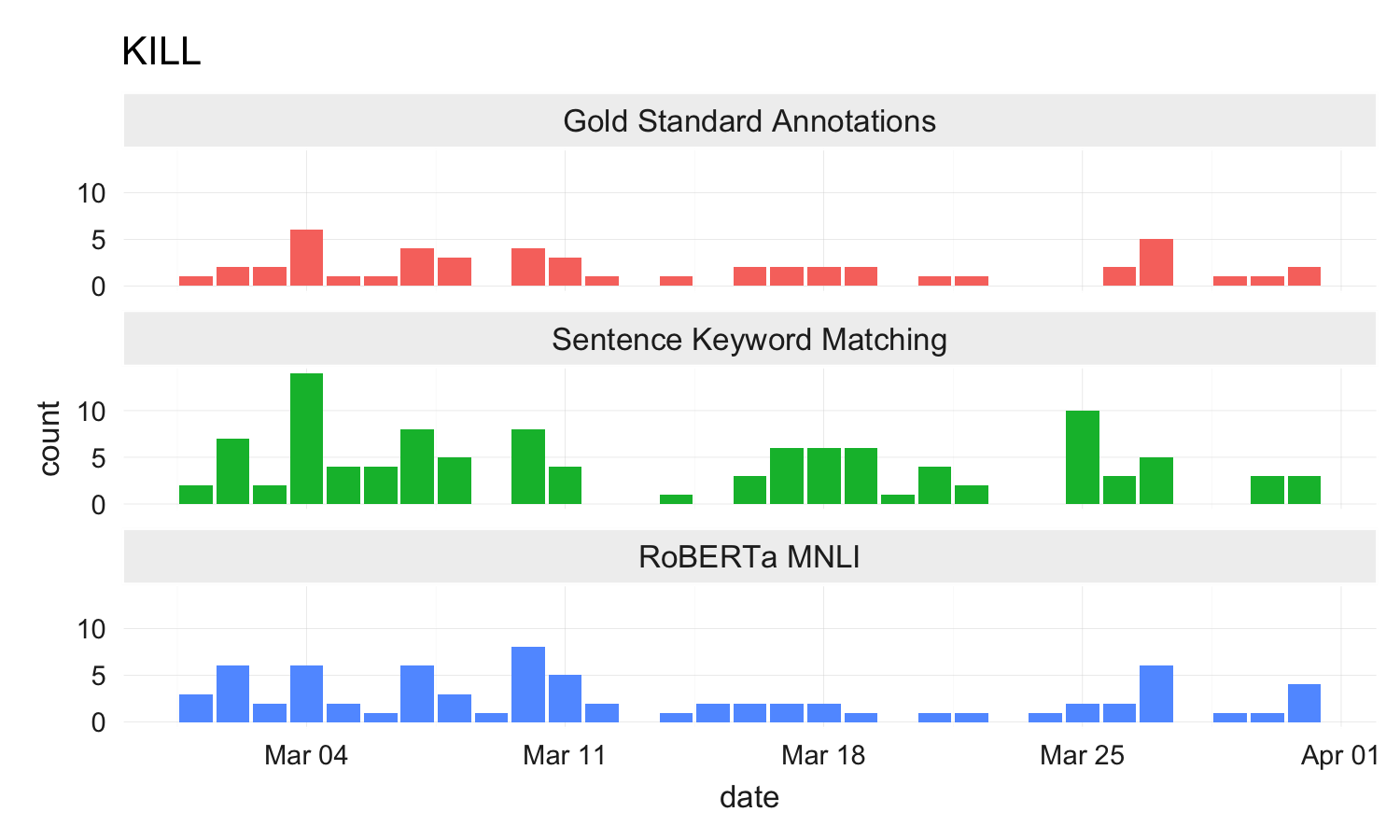}
    \includegraphics[width=\columnwidth]{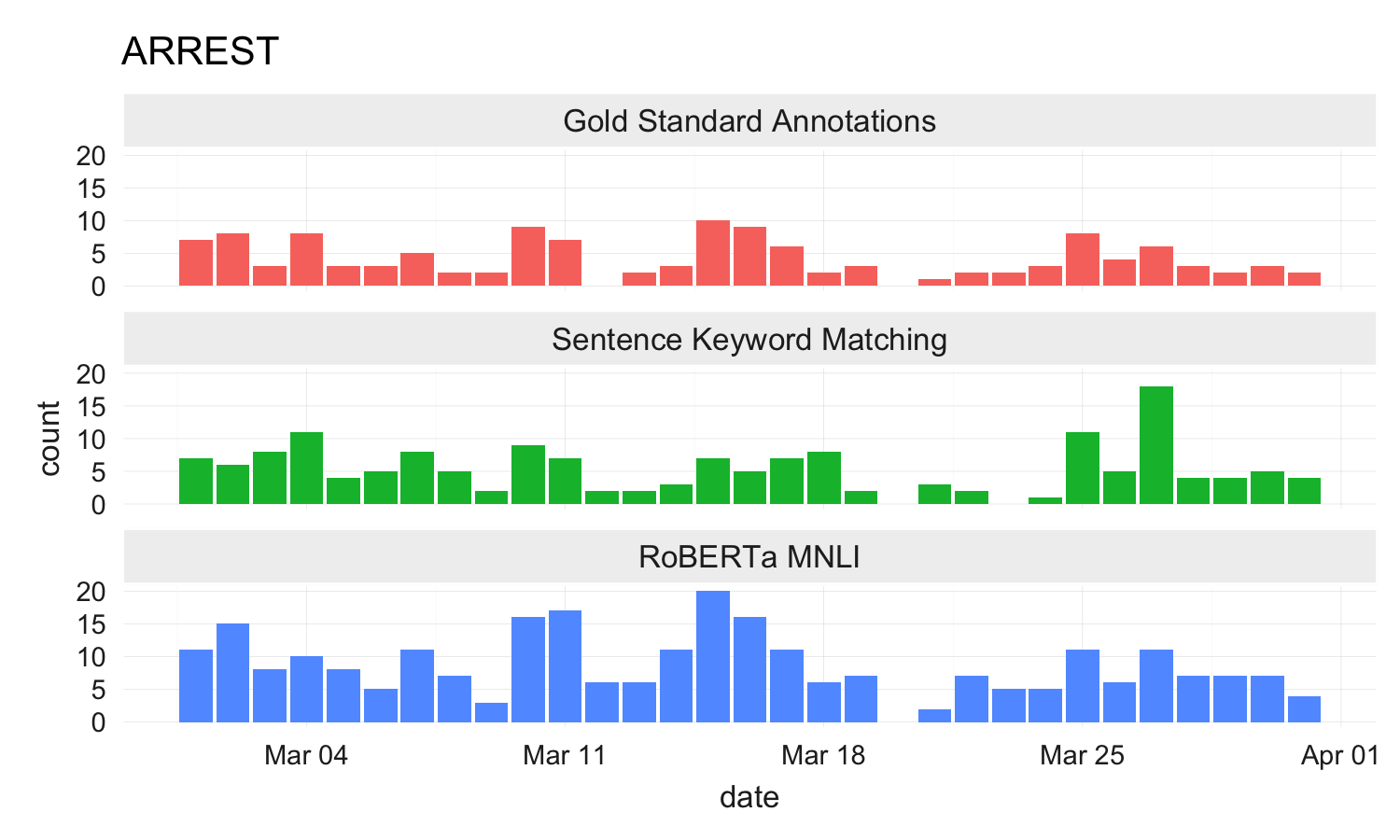}
    \includegraphics[width=\columnwidth]{figs/social_aggregate_fail.png}
    \includegraphics[width=\columnwidth]{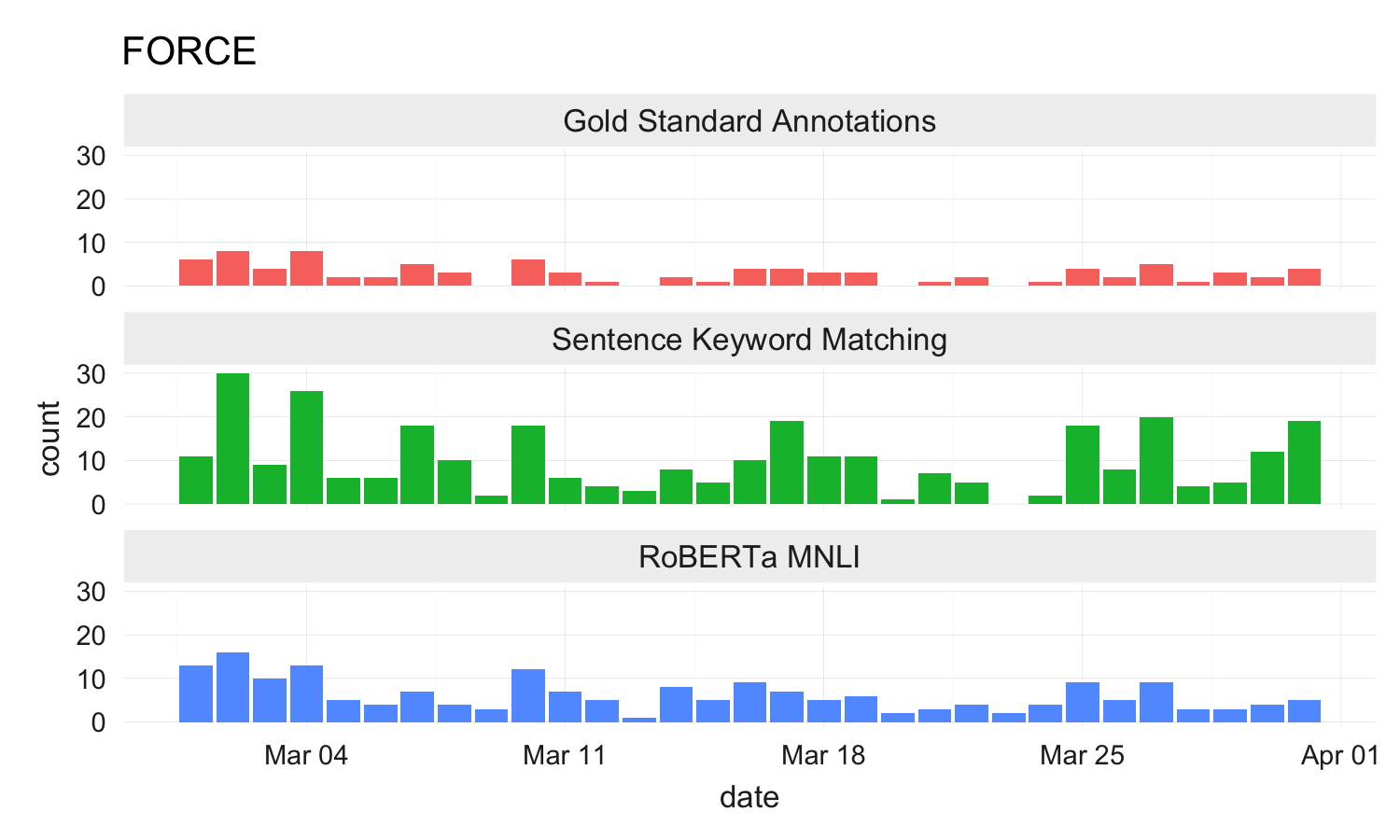}
    \includegraphics[width=\columnwidth]{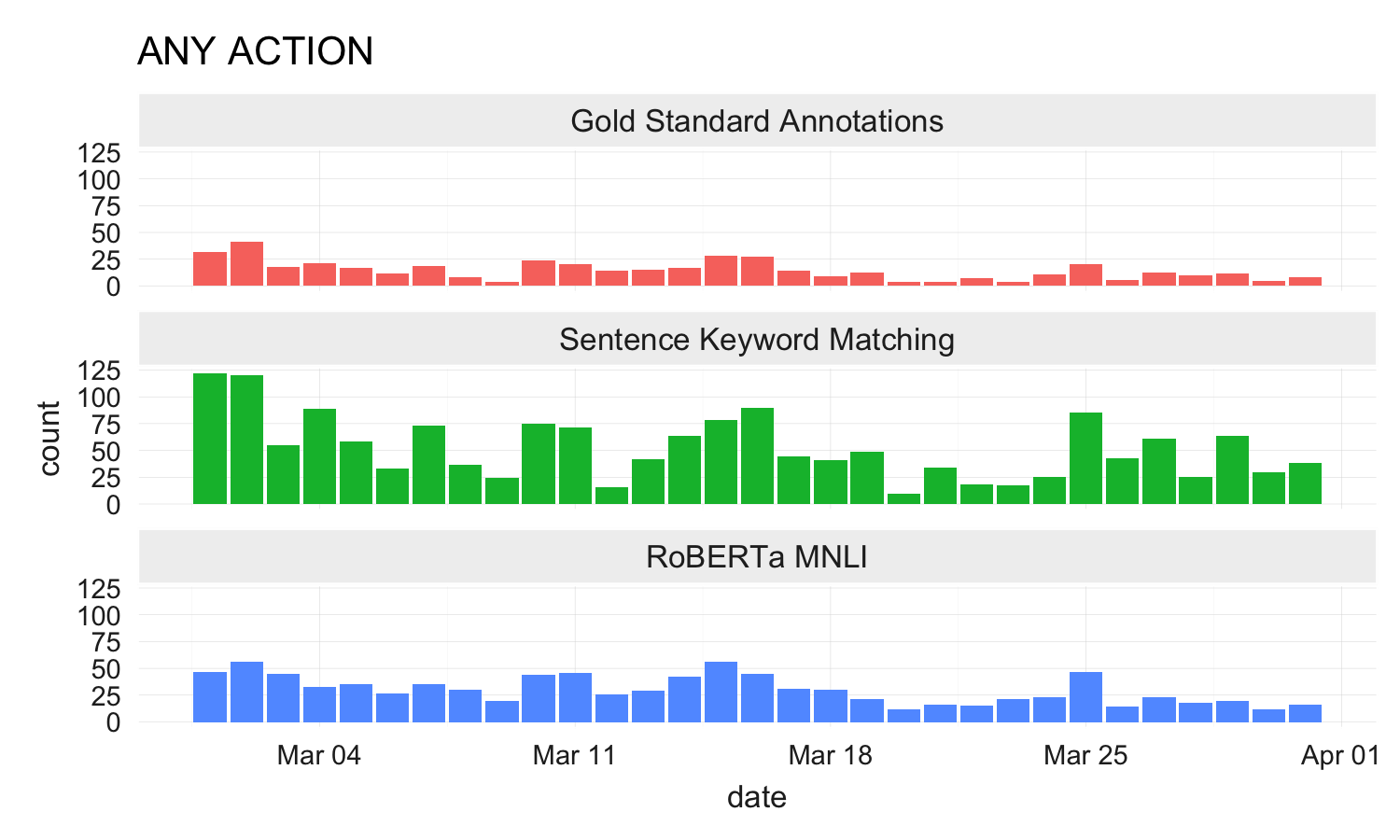}
    \caption{Temporal aggregate figures for all event classes comparing daily document counts of gold standard annotations, sentence keyword matching, and RoBERTa+MNLI.}
    \label{fig:all_aggs}
\end{figure*}

We report the temporal aggregate comparisons for all event classes in Figure \ref{fig:all_aggs} to supplement the figure in the main text showing results for the \failLabel~ class.

\section{Prototype span-based annotation schema} 
Before arriving at the annotations via natural language described in Section~\ref{ss:label-descriptions}, we first attempted to gather \emph{span-based} text annotations in order to collect more fine-grained details about police activity. In these prototype rounds, we first asked annotators to highlight spans in the text that answered ``What action did police do?'' Then given the action text-span they highlighted, we asked them to highlight spans for the following questions: ``Police did the action \emph{using what?''} ``Police did the action \emph{towards whom?}'' ``Where did the action occur?'' ``When did the action occur?'' ``Why did the action occur?'' 

There were several major barriers to this annotation schema that caused us to abandon the span-based annotation approach for our current approach---pre-selecting semantic event classes of interest and having annotators give sentence classification labels. First, we were unable to resolve discrepancies in how much the annotators should highlight for given spans. Following the ``argument reduction criterion'' of \citet{2016stanovskyAnnotating}, we asked annotators to ``highlight as much as you need to answer the question but not more. If you can exclude a word from the highlighting without changing the answer to the question, you should exclude it.'' For example, in the text ``Police suddenly attacked protesters with sticks'' we expected annotators to highlight ``suddenly attacked" versus just ``attacked'' because the former is a slightly different action. However, this criterion did not succeed in improving annotator agreement on span extents. 

Furthermore, with span-based annotations, it was difficult to decide how to properly aggregate police actions (e.g.~how do we automatically separate \emph{suddently attacked} from \emph{attacked} from \emph{did not attack}?) 
Had we been committed to span-based annotations, we may have had to develop much longer, more detailed guidelines, such as those from the Richer Event Description project \cite{o2016richer}. We believe this approach---which requires more work in developing guidelines and training annotators in them---is less easily extensible to new problems and social science domains. Finally, in a training round, the action text spans that annotators did select were not very substantively interesting and worth the additional cost and effort on the part of annotators.\footnote{
Substantively less-interesting police actions identified include \emph{made, identify, placed, recorded, said, spotted, suggested, used}.}

\end{document}